\documentclass[a4paper,fleqn]{cas-dc}

\usepackage[numbers]{natbib}
\usepackage{graphicx}
\usepackage{rotating}
\usepackage[lined,boxed,commentsnumbered]{algorithm2e}
\usepackage{blindtext}
\usepackage{color, soul, framed}
\usepackage{makecell}
\usepackage[numbers]{natbib}
\usepackage{algorithmic} 
\usepackage{hyperref}
\usepackage{color}
\usepackage{blindtext}
\usepackage{color, soul, framed}
\usepackage{url}
\usepackage{hyperref}

\def\tsc#1{\csdef{#1}{\textsc{\lowercase{#1}}\xspace}}
\tsc{WGM}
\tsc{QE}
\tsc{EP}
\tsc{PMS}
\tsc{BEC}
\tsc{DE}

\begin{document}
	\begin{sloppypar}
		\let\WriteBookmarks\relax
		\def\floatpagepagefraction{1}
		\def\textpagefraction{.001}
		\shorttitle{FAIR1M Dataset}
		\shortauthors{Xian Sun et~al.}
		
		\title [mode = title]{FAIR1M: A Benchmark Dataset for Fine-grained Object Recognition in High-Resolution Remote Sensing Imagery}                      
		
		
		
		\author[1,2,3]{Xian Sun}[orcid=0000-0002-0038-9816]
		\ead{sunxian@aircas.ac.cn}
		
		\address[1]{Aerospace Information Research
			Institute, Chinese Academy of Sciences, Beijing 100190, China}

		\author[1,2,3]{Peijin Wang}
		\ead{wangpj@aircas.ac.cn}
		
		\author[1,2,3]{Zhiyuan Yan}
		\ead{yanzy@aircas.ac.cn}

		\author[4]{Feng Xu}
		\ead{fengxu@fudan.edu.cn}

		\author[5]{Ruiping Wang}
		\ead{wangruiping@ict.ac.cn}

		\author[1,2,3]{Wenhui Diao}
		\ead{diaowh@aircas.ac.cn}

		\author[6]{Jin Chen}
		\ead{chenjin_wonder@hotmail.com}

		\author[1,2,3]{Jihao Li}
		\ead{lijihao17@mails.ucas.edu.cn}
		
		\author[1,2,3]{Yingchao Feng}
		\ead{fengyingchao17@mails.ucas.edu.cn}
		
		\author[1,2,3]{Tao Xu}
		\ead{xutao17@mails.ucas.edu.cn}
		
		\author[7]{Martin Weinmann}
		\ead{martin.weinmann@kit.edu}

		\author[7]{Stefan Hinz}
		\ead{stefan.hinz@kit.edu}
		
		\author[8,9]{Cheng Wang}
		\cormark[1]
		\ead{cwang@xmu.edu.cn}
		
		\author[1,2,3]{Kun Fu}
		\cormark[1]
		\ead{fukun@mail.ie.ac.cn}

		\address[2]{School of Electronic, Electrical and Communication Engineering, University of Chinese Academy of Sciences, Beijing 100049, China}

		\address[3]{Key Laboratory of Network Information System Technology (NIST),
			Aerospace Information Research Institute, Chinese Academy of Sciences,
			Beijing 100190, China}

		\address[4]{Key Laboratory for Information Science of Electromagnetic Waves (MoE), Fudan University, Shanghai, China}
		
		\address[5]{Institute of Computing Technology, Chinese Academy of Sciences, Beijing, 100190, China}

		\address[6]{Beijing Remote Sensing Information Institute, Beijing 100011, China}
		
		\address[7]{Institute of Photogrammetry and Remote Sensing, Karlsruhe Institute of Technology,
			Karlsruhe, Germany}
		
		\address[8]{Fujian Key Laboratory of Sensing and Computing for Smart Cities, School of Information Science and Engineering, Xiamen University, Xiamen 361005, China}
		
		\address[9]{Fujian Collaborative Innovation Center for Big Data Applications in Governments, Fuzhou 350003, China}
		
		\cortext[cor1]{Corresponding author}
		
		
		
		
		
		
		
		
		
		
		
		\begin{abstract}
			With the rapid development of deep learning, many deep learning-based approaches have made great achievements in object detection task. It is generally known that deep learning is a data-driven method. Data directly impact the performance of object detectors to some extent. Although existing datasets have included common objects in remote sensing images, they still have some limitations in terms of scale, categories, and images. Therefore, there is a strong requirement for establishing a large-scale benchmark on object detection in high-resolution remote sensing images. In this paper, we propose a novel benchmark dataset with more than 1 million instances and more than 15,000 images for Fine-grAined object recognItion in high-Resolution remote sensing imagery which is named as \textbf{FAIR1M}. We collected remote sensing images with a resolution of 0.3m to 0.8m from different platforms, which are spread across many countries and regions. All objects in the FAIR1M dataset are annotated with respect to 5 categories and 37 sub-categories by oriented bounding boxes. Compared with existing detection datasets dedicated to object detection, the FAIR1M dataset has 4 particular characteristics: (1) it is much larger than other existing object detection datasets both in terms of the quantity of instances and the quantity of images, (2) it provides more rich fine-grained category information for objects in remote sensing images, (3) it contains geographic information such as latitude, longitude and resolution, (4) it provides better image quality owing to a careful data cleaning procedure. To establish a baseline for fine-grained object recognition, we propose a novel evaluation method and benchmark fine-grained object detection tasks and a visual classification task using several State-Of-The-Art deep learning-based models on our FAIR1M dataset. Experimental results strongly indicate that the FAIR1M dataset is closer to practical application and it is considerably more challenging than existing datasets.  
			%
		\end{abstract}
		
		
		
		
		\maketitle
		\section{Introduction}
		
		Object detection and recognition aims to obtain the localization and categories of objects of pre-defined categories in an image. It is one of the most fundamental and important tasks in the field of earth observation, which serves various civil applications, such as geographic information system mapping, agriculture, traffic planning, and navigation \cite{wang2019fmssd,wang2018multiscale,han2015object,sun2021pbnet, sun2020sraf, cheng2016learning,cheng2016survey,deng2018multi}. Due to the wide spatial coverage of remote sensing images, there are typically a large number of objects in a remote sensing image. It is a challenging task for machines to recognize and detect objects accurately in such images, but the development of deep learning-based approaches provides effective solutions characterized by a strong ability of feature extraction and feature expression.

		
		

		However, deep learning is a data-driven concept in the field of computer vision, and the performance of respective deep learning-based approaches strongly depends on the quality and quantity of given data. A challenging and excellent dataset can accelerate the development of the field. For example, the ImageNet \cite{deng2009imagenet} and MSCOCO \cite{lin2014microsoft} datasets hasten the evolution of Convolutional Neural Networks (CNNs) on natural scene image classification and object detection tasks, UC Merced \cite{yang2010bag} and MSTAR\cite{diemunsch1998moving} datasets separately promote the progress of optical remote sensing scene classification and Synthetic Aperture Radar (SAR) target recognition, the Cityscape \cite{Cordts2016Cityscapes} and Vaihingen \cite{Rottensteiner_et_al_2012} datasets facilitate the development of deep neural networks for semantic segmentation in natural scenes and remote sensing scenes respectively, the DOTA \cite{xia2018dota} and DIOR \cite{li2020object} datasets are proposed for generic object detection in remote sensing images, and the FGSD \cite{chen2020fgsd} and VEDAI \cite{razakarivony2016vehicle} datasets inspire the research of fine-grained object detection in remote sensing images. Among these tasks, object detection and recognition has attracted wide attention in the past few years \cite{han2017an,guo2018geospatial,xu2017deformable,wu2019orsim,zhang2019hierarchical,chen2014vehicle,yan2019iou-adaptive}. However, compared with datasets focusing on natural scenes, some deficiencies of existing datasets limit the development of fine-grained object recognition in the field of remote sensing.
		
		\begin{figure}[t]
			\centering
			\includegraphics[width=1.\linewidth]{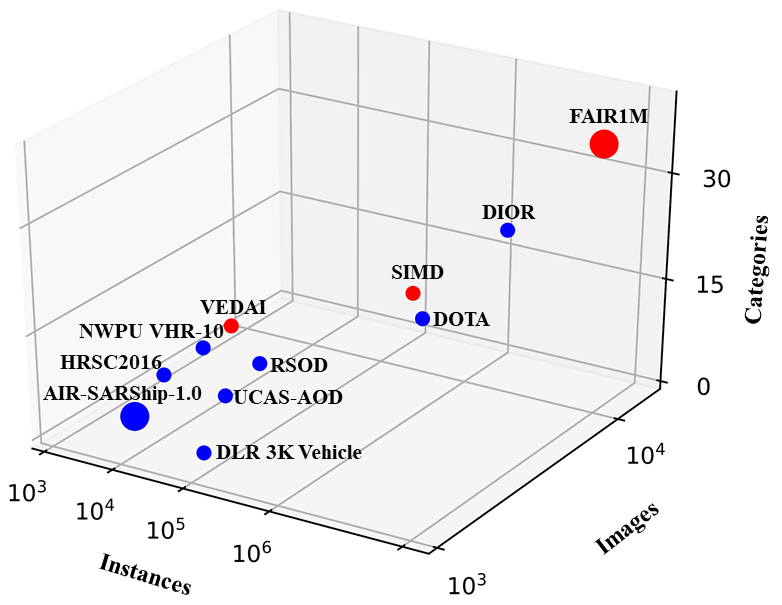}
			\caption{Multi-dimensional representation for typical object detection datasets in the field of remote sensing. Red and blue points denote the fine-grained datasets and generic datasets, respectively. The larger points denote the datasets which additionally contain geographic information. }
			\label{FIG:3D}
		\end{figure}

		\begin{figure*}
			\centering 
			
			\includegraphics[width=16.4cm]{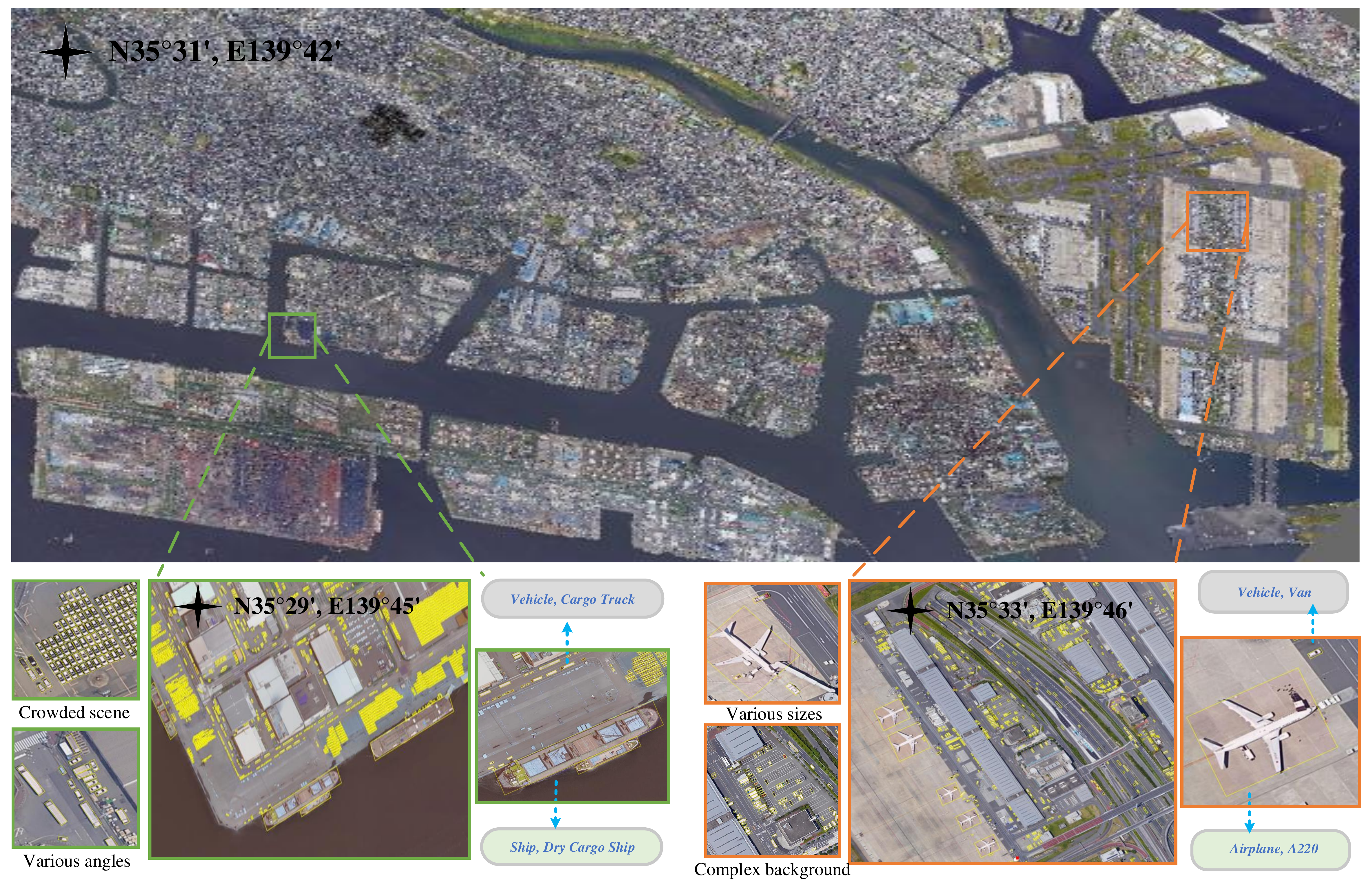}

			\caption{Visualization of annotations in the FAIR1M dataset. In addition to fine-grained object categories, the FAIR1M dataset contains crowded scenes, complex background, and various sizes and angles of objects. Yellow boxes are the bounding boxes we annotated.}
			\label{Figure: Representative}
		\end{figure*}
		

		
		\textbf{1. The scale of datasets can be further expanded.} With the enhancement of the demand for remote sensing applications, object detectors need to own stronger generalization ability. Due to the over-fitting phenomenon, some seemingly excellent algorithms which perform well on small datasets are likely to obtain bad results in a larger dataset. Hence, in order to evaluate an algorithm more comprehensively, the scale of corresponding datasets is required to be relatively large in terms of object instances and image quantity. Currently, noteworthy achievements have been made in natural scene object detection datasets, such as MSCOCO \cite{lin2014microsoft}. Therefore, the scale of remote sensing scene object detection datasets still needs to be expanded eagerly.
		
		\textbf{2. Fine-grained information needs to enrich.} Objects in remote sensing images usually have multiple fine-grained types. At present, generic object detection and recognition are difficult to meet the needs of applications and the demand for fine-grained recognition is rapidly growing. For example, an excellent algorithm must be able to not only correctly detect instances belonging to the category \textit{Airplane}, but also recognize that the object belongs to a certain sub-category such as \textit{Airbus 350}, \textit{Boeing 747} or other type. As shown in Figure \ref{FIG:3D}, most well-known existing object detection datasets contain coarse-grained annotation information, or a small amount of fine-grained information. For instance, DOTA \cite{xia2018dota} divides vehicles into \textit{Large-vehicles} and \textit{Small-vehicles}. Deep learning models trained on these datasets may not perform very well when they are faced with large-scale type recognition tasks. 
		
		\textbf{3. Image quality needs to be improved.} Taking into account the process of high-resolution satellite image acquisition, there may be some interference factors in the images, such as clouds and fog. No cleaning or improper cleaning will directly reduce the quality of images, and then influence the performance of object detection algorithms. 
		
		\textbf{4. The dataset should contain more geographic characteristics.} Temporal and spatial information are the two major geographic characteristics in the field of remote sensing. Remote sensing images collected in most of the existing object detection datasets are single-temporal, which means that there is no time dimension difference in the same remote sensing scene. It is relatively difficult for these datasets to represent the change of seasons and surroundings. This also has an effect on the generalization ability of deep learning models to some extent. As shown in Figure \ref{FIG:3D}, the storage format of images in existing datasets is the same as for natural scene images and tends to lack geographic information. Geographic information in turn is vital for remote sensing image processing, referring to properties such as spatial resolution, longitude, and latitude.

		Consequently, in order to better address the problems mentioned above, we propose a novel benchmark dataset for Fine-grAined object recognItion in high-Resolution remote sensing imagery which is named as \textbf{FAIR1M}. Some representative examples of images and their annotations are shown in Figure \ref{Figure: Representative}. 
		In the FAIR1M dataset, we collect remote sensing images containing more than 15,000 images and 1 million instances from the Gaofen satellites and Google Earth platform. Scenes in these remote sensing images are spread across many continents, such as Asia, America, and Europe. All FAIR1M images are annotated with oriented bounding boxes (OBB) and with respect to 5 categories and 37 sub-categories under the guidance of many experts in remote sensing. To the best of our knowledge, FAIR1M is the largest fine-grained oriented object detection dataset suitable for remote sensing scenes.
		
		Due to the high quantity and quality of images and fine-grained categories, FAIR1M promotes the challenging tasks for fine-grained object detection and visual classification, which aims to obtain fine-grained categories and locations of objects. Compared with generic object detection, fine-grained object detection can recognize not only generic categories but also fine-grained types for objects in remote sensing images. In order to standardize the development of fine-grained detection on the FAIR1M dataset, we also propose a new evaluation metric and build a benchmark using representative algorithms.
		
		
		
		
		
		In summary, the proposed FAIR1M benchmark dataset intends to provide a large-scale fine-grained object recognition dataset to the remote sensing community. We hope that with the help of FAIR1M, a growing number of novel algorithms can be investigated in the field of remote sensing image interpretation. Our main contributions of this work are briefly summarized as follows:
		\begin{itemize} 
			\item A large-scale public dataset has been proposed for object recognition in remote sensing images. To the best of our knowledge, the proposed FAIR1M benchmark dataset is the largest fine-grained object recognition dataset in remote sensing with more than 1 million instances. Moreover, multi-temporal images, geographic information and orientated annotations are provided in the FAIR1M dataset.  
			\item To evaluate the performance of detection methods on the FAIR1M dataset, we design a novel evaluation metric for fine-grained object detection and recognition in remote sensing imagery. Compared with generic object detection, fine-grained object detection pays more attention to the categories of objects. As a result, a score-aware and challenging evaluation metric is proposed and validated on the FAIR1M dataset.
			\item We propose a novel cascaded hierarchical object detection network and benchmark fine-grained object detection tasks and a visual classification task using several state-of-the-art object detection models on the FAIR1M dataset, which can be utilized as the baseline for future work. We believe that this benchmark will certainly promote the development of fine-grained object detection and recognition in the field of remote sensing. Furthermore, the FAIR1M dataset will be released as the ISPRS benchmark dataset on the benchmark website\footnote{http://gaofen-challenge.com/.}.
		\end{itemize}


		The rest of this paper is organized as follows. In Section 2, we review several existing popular object detection datasets in remote sensing. Section 3 describes the proposed new FAIR1M dataset in detail. In Section 4, we introduce the evaluation metrics and evaluate some excellent object detectors on the proposed FAIR1M dataset. Finally, Section 5 summarizes the paper and provides an outlook on the future of object detection in remote sensing imagery.
		
		
		
		
		\begin{table*}[t]
			\renewcommand{\arraystretch}{1.4}
			\setlength{\tabcolsep}{2.7pt}
			\caption{Comparison between FAIR1M dataset and other object detection datasets containing remote sensing images}
			\label{table_dataset}
			\label{tbl1}
			\begin{tabular}{|c|p{2.cm}|c|c|c|c|c|c|c|c|}
				\toprule
				Datasets		& \qquad Source		& Instances		& Images	& Image width & Categories	& Annotation & Image format	& Fine-grained	\\
				\midrule
				\midrule
				NWPU VHR-10	\cite{cheng2016learning}	& Google Earth		& 3,775			& 800	&$\sim$1000	& 10			& HBB	& JPG		& N	\\		
				VEDAI \cite{razakarivony2016vehicle}			& Google Earth		& 3,640			& 1,210	& 512, 1024	& 9				& OBB	& PNG		& Y	\\
				UCAS-AOD \cite{zhu2015orientation}		& Google Earth		& 6,029			& 910	&$\sim$1000	& 2				& OBB		& PNG	& N	\\
				DLR 3K Vehicle \cite{liu2015fast}	& Aerial images 		& 14,235		& 20	&5616	& 2				& OBB	& JPG		& N	\\
				HRSC2016 \cite{liu2016ship}		& Google Earth		& 2,976			& 1,070	&$\sim$1100	& 1				& OBB		& BMP	& N	\\
				AIR-SARShip-1.0 \cite{xian2019air}	& Gaofen-3		& 3,000			& 31	&3000	& 1				& HBB		& TIFF	& N	\\
				RSOD \cite{xiao2015elliptic}			& Google Earth, Tianditu		& 6,950			& 976	&$\sim$1000	& 4				& HBB		& JPG	& N	\\
				DOTA \cite{xia2018dota}			& Google Earth, Satellite JL-1, GF-2		& 188,282		& 2,806	&800-4000	& 15 			& OBB	& PNG		& N	\\
				xView \cite{lam2018xview}			& WorldView-3		& 1 million 	& 1,127	&2000-4000	& 60			& HBB	& PNG		& Y	\\
				DIOR \cite{li2020object}			& Google Earth		& 192,472		& 23,463 &800	& 20 			& HBB & JPG			& N	\\
				SIMD \cite{haroon2020multi}			& Google Earth		& 45,096 		& 5,000 &1024	& 15 			& HBB		& JPG	& Y	\\
				FGSD \cite{chen2020fgsd}			& Google Earth		& 5,634 		& 2,612 &930	& 43 			& OBB		& JPG	& Y	\\
				\midrule
				FAIR1M 			&  Gaofen, Google Earth 		& 	 \textbf{1.02 million}			& 15,266		&1000-10000	& 37 			& OBB	& TIFF		& Y	\\
				\bottomrule
			\end{tabular}
			
		\end{table*}


		\begin{figure*}
			\centering
			\includegraphics[scale=0.25]{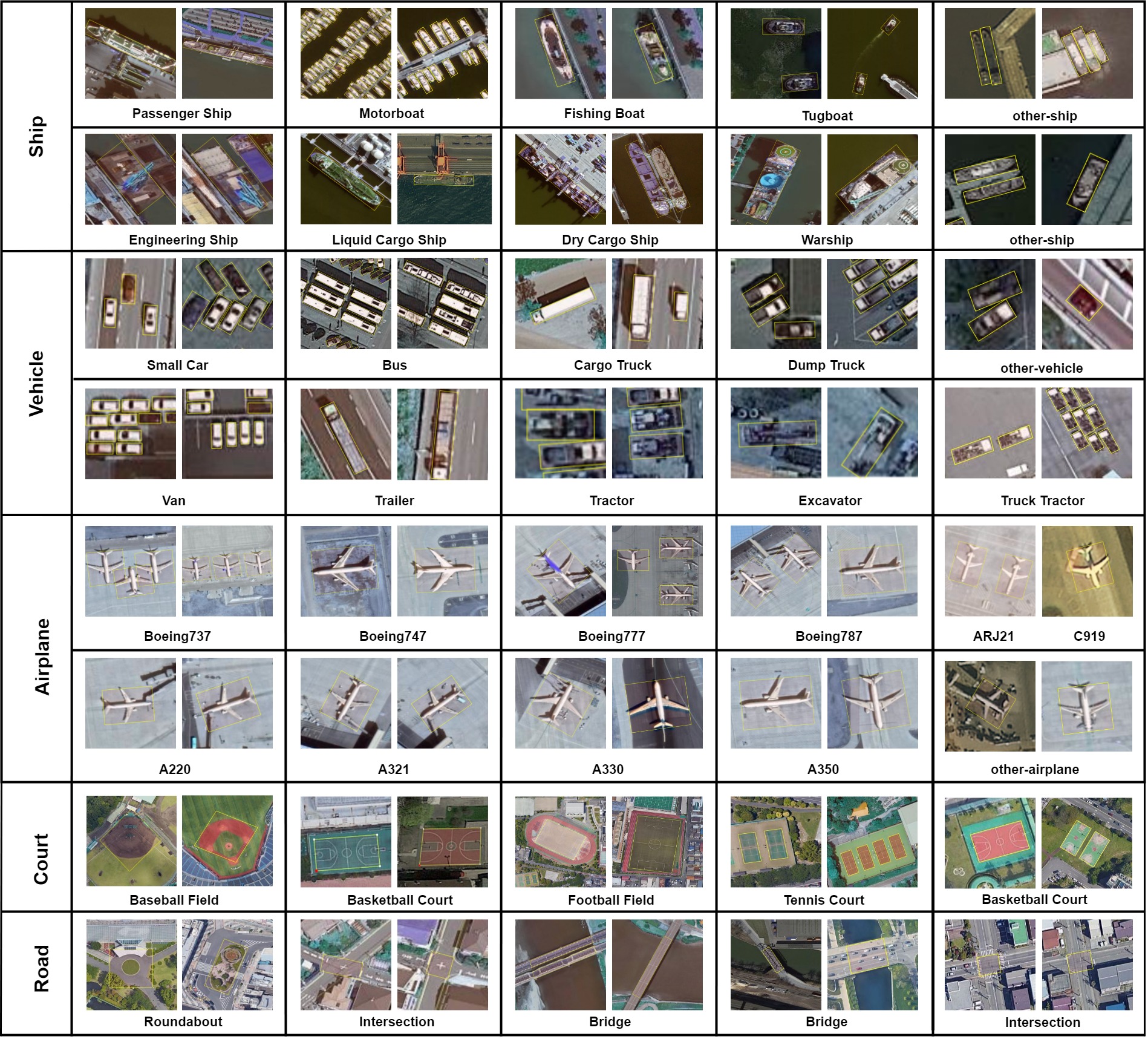}
			\caption{Data samples of each category in the FAIR1M dataset. }
			\label{FIG:class}
		\end{figure*}
		
		
		


		\section{Related Work}
		
		\subsection{Datasets for Object Recognition in Ground-level Natural Images}
		
		Object detection is one of the most important techniques in the fields of computer vision and remote sensing. Therefore, a series of datasets were proposed to accelerate the development of object detection methods. The PASCAL VOC \cite{everingham2010pascal} is one of the fundamental datasets, which is widely used in the ground-level natural object detection. It was proposed in 2005 and has been gradually extended. The Pascal VOC dataset contains 20 object categories in 11.5k images with 27k bounding boxes. This dataset is widely used by researchers. However, the number of categories and the number of instances limit the development of methods based on this dataset. The MSCOCO \cite{lin2014microsoft} dataset proposed in 2014 is much larger than the PASCAL VOC dataset, which contains more object categories and object instances, especially the objects with smaller size. Specifically, the dataset contains 80 object categories and 896k object instances in more than 200k images.
		
		More recently, Megvii released Objects365 \cite{shao2019objects365}, a large-scale dataset containing 638k images, which is 5 times larger than MSCOCO dataset. It consists of 365 object classes and more than 10 million object instances. Compared with the above datasets, the number of instances per image reaches 15.8 (2.4 for PASCAL VOC and 7.3 for MSCOCO). Furthermore, Google proposed the Open Images \cite{kuznetsova2018open} dataset, which contains 16 million bounding boxes, 600 categories, and 1.9 million images. As a result, it is the largest existing dataset dedicated to object detection. The above large-scale datasets promote the development of algorithms on general object detection in natural images \cite{kong2016hypernet:,huang2017speed/accuracy,shen2017dsod:,li2017fssd:,xu2017deformable,ren2018deformable,redmon2018yolov3}. However, considering the differences between the ground-level natural images and geo-spatial remote sensing images, it is necessary to specially design geo-spatial object detection dataset for the development of remote sensing.

		\subsection{Datasets for Object Recognition in Remote Sensing Images}
		
		Due to the top view of the remote sensing images and the various spatial resolutions of sensors, the scale variations of object instances are huge and objects often appear in arbitrary orientations. In order to promote the development of the aerial object detection research, many remote sensing datasets have been proposed, such as NWPU VHR-10, HRRSD, DOTA and DIOR. The NWPU VHR-10 dataset \cite{cheng2016learning} is composed of 10 geospatial object categories in 715 images, and the images are collected from Google Earth. However, the total number of object instances in NWPU VHR-10 is only 3775, which cannot adequately reflect the complexity of the problem in the real world. The HRRSD dataset \cite{zhang2019hierarchical} contains more than 55k object instances for 13 object categories. The images in the HRRSD dataset have been acquired from Google Earth and Baidu Map and their spatial resolution varies from 0.15m to 1.2m. However, the size of the images in the HRRSD dataset is relatively small, with only 227 $\times$ 227 pixels, which limits the range of the research. Recently, a large-scale dataset has been proposed, named DIOR dataset \cite{li2020object}, which contains more than 23k images and 192k instances, covering 20 object categories. However, the objects in the DIOR dataset are annotated with horizontal bounding box (HBB), which cannot better enclose the objects and solve the problem of oriented object detection. The DOTA dataset \cite{xia2018dota} contains 15 different geospatial object categories and more than 188k oriented bounding boxes. The dataset consists of 2806 images collected from different platforms with multiple spatial resolutions and the sizes of images range from 800 $\times$ 800 to 4000 $\times$ 4000 pixels, which is widely used in the research of object detection.
		
		There are some other remote sensing datasets dedicated to the research on important object categories. The RSOD dataset \cite{xiao2015elliptic} was proposed in 2015, which contains four categories, including overpasses, oil tanks, airplanes and playgrounds. It consists of 976 images obtained from Google Earth and Tianditu with spatial resolutions ranging from 0.3m to 3m. However, the RSOD dataset only has 6,950 instances. The UCAS-AOD dataset \cite{zhu2015orientation} is designed for airplane and vehicle detection, which consists of an airplane dataset and a vehicle dataset. The former contains 600 images and 3210 airplanes and the latter contains 310 images and 2819 vehicles. The COWC dataset \cite{mundhenk2016large} only has one object category. It is designed for car detection, which contains about 32.7k instances. The LEVIR dataset \cite{zou2017random} contains three object classes, including airplane, ship and oilpot. It consists of 21.9k images and 11k object bounding boxes. The size of the images is 600 $\times$ 800 pixels and the images have been collected from Google Earth and their spatial resolutions vary from 0.2m to 1m. There are also datasets dedicated to building detection, such as the Semcity toulouse dataset \cite{roscher2020semcity}, ISPRS benchmark on urban object detection and 3D building reconstruction dataset \cite{rottensteiner2014results}, Inria Aerial Image Labeling dataset \cite{maggiori2017can} and DeepGlobe 2018 dataset \cite{demir2018deepglobe}. However, these datasets ignore the fine-grained category information for such important geo-spatial objects.

		\subsection{Datasets for Fine-grained Objects Detection and Recognition}
		
		Fine-grained object recognition is a significant task in the field of remote sensing. Therefore, there are some datasets have been proposed. The VEDAI dataset \cite{razakarivony2016vehicle} is a fine-grained vehicle detection dataset, which promotes the development of fine-grained vehicle detection algorithms in remote sensing images. There are in total 1210 images and 3700 instances in VEDAI and the size of the images is 1024 $\times$ 1024 pixels. It can be seen from the number of instances and images that the distribution of instances in this dataset is relatively sparse. The MTARSI dataset \cite{wu2020benchmark} is designed for the airplane type recognition, which consists of 20 airplane types. A total of 9385 remote sensing images were obtained from Google Earth satellite images, with spatial resolution ranging from 0.3m to 1.0m. However, the size of the images in MTARSI is only 256 $\times$ 256. The MTARSI dataset only has the annotation of category and cannot be used for the object detection task. The HRSC2016 dataset \cite{liu2017a} is the most used dataset for ship detection in remote sensing images, which contains 22 classes of ships but only has 1061 images. The image resolutions are between 0.4m and 2m and the image sizes are ranging from 300 $\times$ 300 to 1500 $\times$ 900 pixels. However, the images in HRSC2016 only cover a very limited number of ports and the number of ships are limited. The FGSD dataset \cite{chen2020fgsd} is the latest ship detection dataset, which consists of high-resolution satellite images from 17 large ports in four countries and 43 classes of ships. It contains 5634 instances and 2612 images. However, the number of images and instances still limits its application. Therefore, it is significant to propose a standard and large-scale dataset for applying deep learning-based methods to the field of remote sensing.
		


		\begin{figure*}
			\centering
			\includegraphics[scale=0.9]{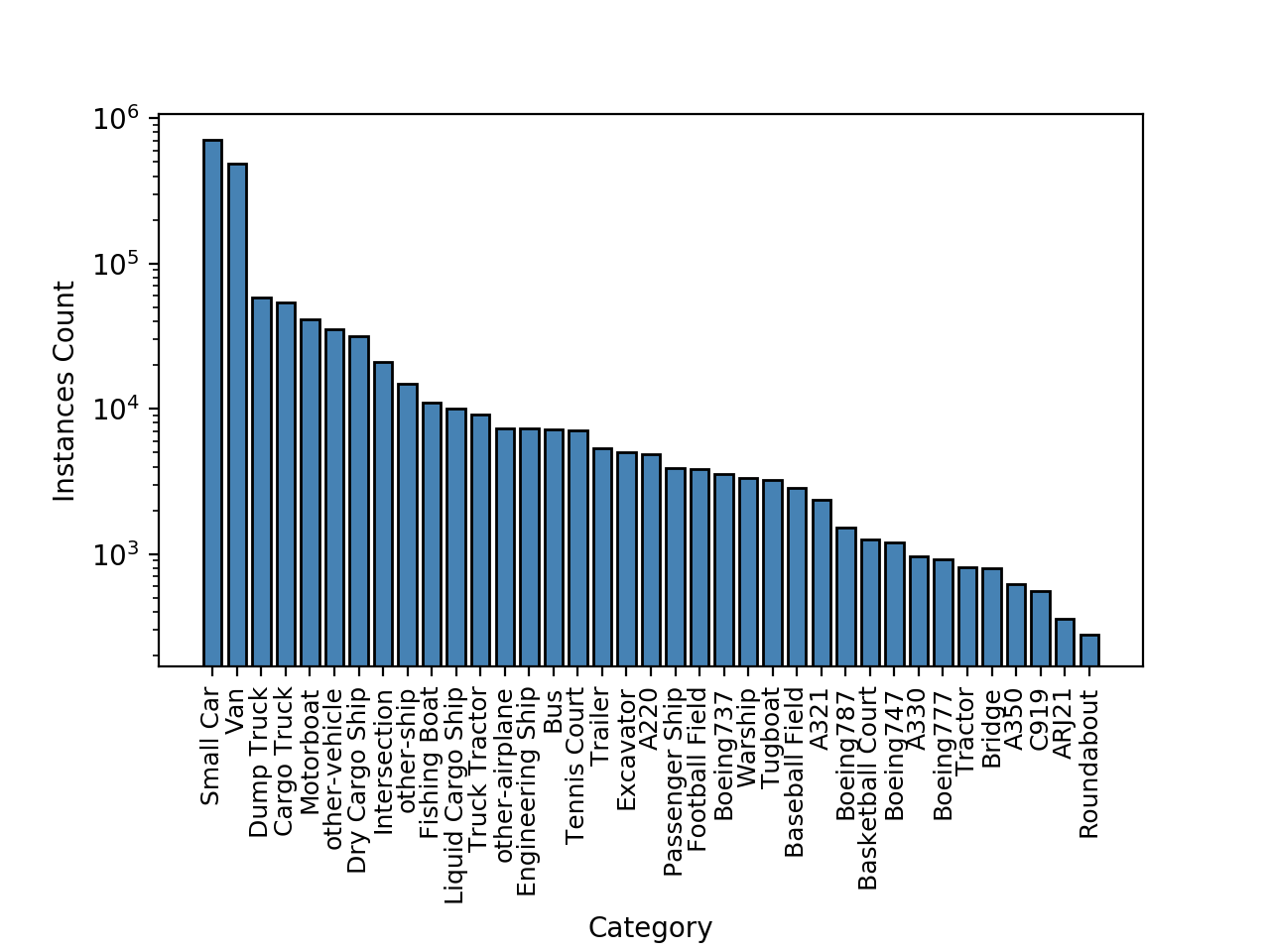}
			\caption{The distribution of the number of instances per category. }
			\label{FIG:number2}
		\end{figure*}

		\begin{figure*}
			\begin{minipage}{0.49\linewidth}
				\centerline{\includegraphics[width=1\linewidth]{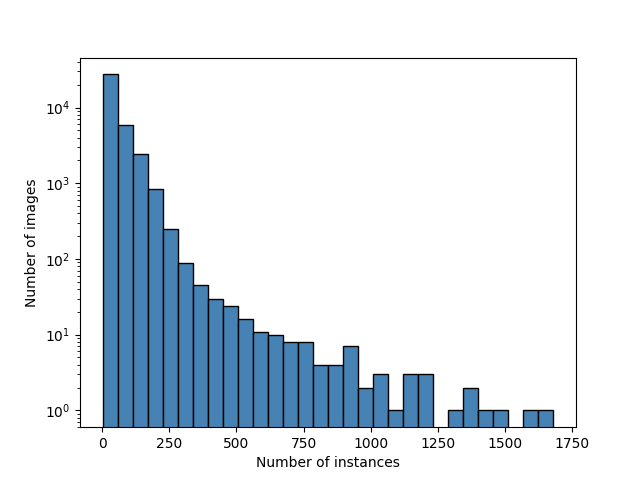}}
				\centerline{(a)}
			\end{minipage}
			\hfill
			\begin{minipage}{0.49\linewidth}
				\centerline{\includegraphics[width=1\linewidth]{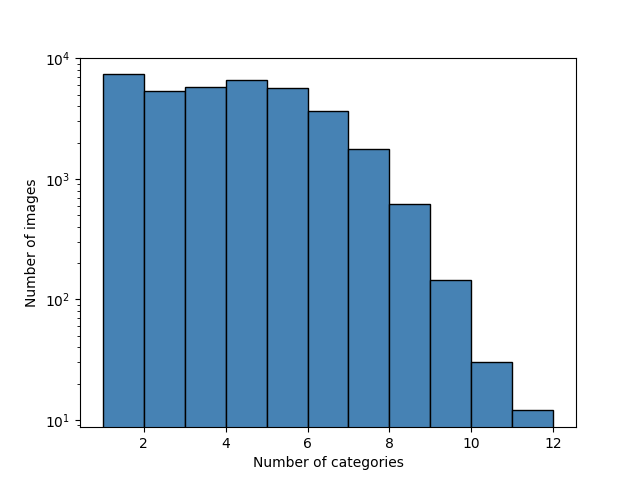}}
				\centerline{(b)}
			\end{minipage}
			
			\caption{{\color{black}{(a) The distribution of the number of instances per image. (b) The distribution of the number of categories per image.}}}
			\label{Figure:number}
		\end{figure*}
		
		Above datasets are designed for fine-grained single-class object detection and recognition. There are two datasets for fine-grained multi-class object detection and recognition, which are the xView dataset \cite{lam2018xview} and SIMD dataset \cite{haroon2020multi}. The xView dataset contains over 1 million horizontal bounding boxes in 1127 images, labeled with respect to 60 object categories, which were organized in a class hierarchy. There are mainly seven categories, including fixed-wing aircraft, passenger cars, trucks, railway vehicles, engineering vehicles, maritime ships, and buildings. However, the image quality of the xView dataset is not very high and some significant categories are coarse, for example, airplanes only contain two types (i.e., Small Aircraft and Cargo Plane). Besides, the distribution of object instances in the xView dataset is relatively unbalanced, most of which focus on the building class and the small car class. The SIMD dataset contains 15 different object categories, including seven types of vehicles, six types of aircrafts, Boat and Others class. The images were acquired from Google Earth. Specifically, the dataset comprises 5000 images of resolution 1024 $\times$ 768 pixels and collectively contains around 45k objects. However, the SIMD dataset also applies the horizontal bounding box definition for object annotation and the distribution of object instances in SIMD is also unbalanced. Half of the instances belong to the Car class, there are only few instances in the other categories. The FGSD dataset consists of 43 fine-grained categories, but it only contains ships. Our dataset applies the oriented bounding box definition for annotation and focuses on a careful selection of fine-grained categories for significant objects in remote sensing images, including airplanes, ships, vehicles, courts and roads. It is a larger and more comprehensive dataset for object detection.

		\section{Details of the FAIR1M Dataset }
		\subsection{Image collection and Pre-processing}
\begin{figure}
			\centering
			\includegraphics[scale=0.5]{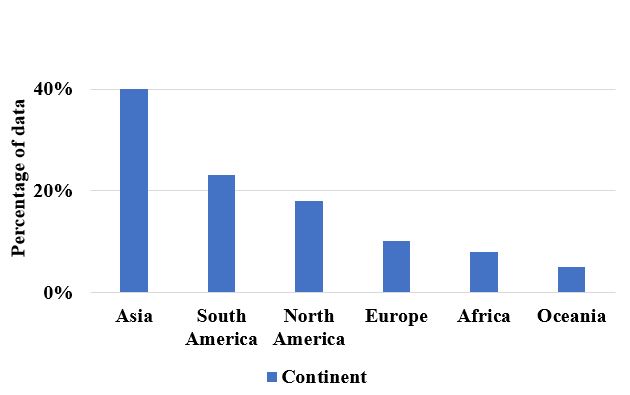}
			\caption{The distribution of the FAIR1M dataset across continents. It can be seen from the figure that the FAIR1M dataset is widely distributed all over the world.}
			\label{FIG:world}
		\end{figure}

		Xia et al. \cite{xia2018dota} prove that a variety of sensors and resolutions can be used to eliminate biases. To meet the needs of practical applications, images in our dataset are collected from the Gaofen satellites and Google Earth, with a spatial resolution ranging from 0.3m to 0.8m. The diversity of data is of great significance to the study of transportation and humanities in different countries and regions. Therefore, we collect more than 15,000 images from more than 100 civil airports, harbors, and cities all over the world. The distribution of the FAIR1M dataset with respect to different continents can be seen in Figure \ref{FIG:world}.

		\begin{figure*}
			\centering
			\includegraphics[scale=1.2]{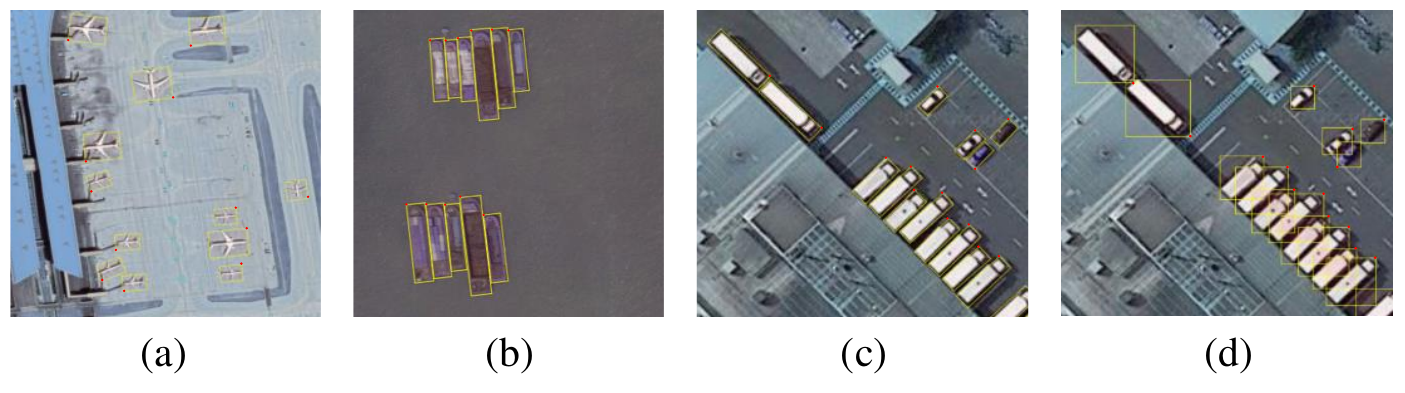}
			
			\caption{Visualization of annotations of FAIR1M dataset. (a), (b) and (c) show the oriented bounding box annotations of airplanes, ships, and vehicles, respectively. The red points represent the top-left points of the instances. (d) shows the traditional horizontal annotation in other datasets.}
			\label{FIG:anno}
		\end{figure*}

		To obtain high-resolution remote sensing images, we adopt a series of pre-processing methods for Gaofen satellite images. Considering the poor quality of satellite images in some datasets, we first check the quality of the raw data and remove the images with many clouds, noise, and bright spots. In order to ensure that the images in the same region have the same positioning accuracy, we perform block adjustment for multi-temporal and multi-source images. Based on the results of block adjustment, we perform a rational function to generate the orthographic results of panchromatic and multi-spectral images. Then, we use the Pan Sharpening algorithms \cite{padwick2010worldview} to improve the spatial resolution of multispectral images by fusing panchromatic images. Finally, we use histogram equalization to adjust the hue component of the images. In this way, we can obtain high-resolution and high-quality remote sensing images.

		\subsection{Category Design}
		
		Most existing datasets pay more attention to static objects, such as bridges, baseball fields, storage tanks, basketball courts and so on. These datasets lack fine-grained information about objects, which plays an important role in real applications in the field of remote sensing. In the FAIR1M dataset, the objects we selected include 5 categories: airplanes, ships, vehicles, courts and roads. The selection of fine-grained types of each category in the FAIR1M dataset depends on practical application scenarios and the shape it presents. For airplanes, we set 10 fine-grained categories covering 34 airports around the world. The types of airplane contain Boeing 737, Boeing 777, Boeing 747, Boeing 787, Airbus A320, Airbus A220, Airbus A330, Airbus A350, COMAC C919, and COMAC ARJ21, which are the most common categories in the civil aviation. The categories of ships are defined according to their functions. There are 8 specific categories for ships, including passenger ship, motorboat, fishing boat, tugboat, engineering ship, liquid cargo ship, dry cargo ship, and warship. As well as for ships, the categories of vehicles are defined according to their functions. There are 9 specific categories for vehicles, including small car, bus, cargo truck, dump truck, van, trailer, tractor, truck tractor, and excavator. Furthermore, we selected 4 categories and 3 categories for courts and roads, respectively. As a result, there are 37 fine-grained categories in the FAIR1M dataset for object detection. Besides these specific categories, we also assigned the categories ’other-airplane’, ’other-ship’, and ’other-vehicle’ for objects that do not belong to the previously defined specific object types. All categories and the number of instances per category can be seen in Figure \ref{FIG:class}, Figure \ref{FIG:number2} and Figure \ref{Figure:number}. It is well-known that the number of each category depends on its actual distribution in remote sensing scenarios. The distribution of instances can reflect the authenticity and challenge of the proposed dataset. 
		
		

		\subsection{Image Annotation}
		\subsubsection{Annotation Format}
		With the development of deep learning, most of the current detection methods in the field of remote sensing are transferred from natural scenes. As a result, most of existing object detection datasets in remote sensing are annotated with horizontal bounding boxes, such as UCAS-AOD \cite{zhu2015orientation}, NWPU VHR-10 \cite{cheng2016learning}, DIOR \cite{li2020object}, and SIMD \cite{haroon2020multi}. Unlike the objects in natural scenes, which are usually in vertical directions, objects in remote sensing images have a variety of directions. Therefore, using horizontal bounding boxes cannot provide accurate spatial information for oriented objects. To wrap objects more accurately and develop more suitable algorithms for oriented objects in remote sensing images, all instances in the FAIR1M dataset are annotated with oriented bounding boxes (OBB). Samples of annotated instances in the FAIR1M dataset can be seen in Figure \ref{FIG:anno}.
		
		Arbitrary rectangular bounding boxes \{$(x_{i}$,$y_{i}$)\}, i = 1, 2, 3, 4\} are adopted for annotation, where ($x_{i}$,$y_{i}$) denotes the coordinates of the i-{th} vertices of the rectangular bounding boxes. As well as given for other oriented detection datasets, we arrange the vertices of a object in a clockwise direction. Further, the top-left point is highlighted as the first point ($x_{1}$,$y_{1}$), which also denotes the positive direction of the object. For categories that are difficult to recognize, we label them as 'other' category. For instances that are occluded or difficult to identify the positive direction, we add the corresponding label information. 
		
		Considering the down-sampling characteristics of deep neural networks, we set 16 pixels to be the thresholds for annotations. For example, ships which are longer than 13 meters need not be annotated in the images with spatial resolution of 0.8 meters. 
		
		\subsubsection{Annotation Quality}
		To ensure the quality of annotations, we develop a complete and strict quality control process. After getting the initial annotations, we design three stages to check and correct the annotations. At the first stage, we divide annotators in pairs and they check the annotations of each other by annotating them again. Then, each annotator needs to merge the two annotations to obtain a more accurate result. Next, we invite a supervisor to further check the quality of annotations, including location, category, and orientation of bounding boxes. Finally, experts in the field of remote sensing imagery are invited to check the quality of the annotations and the dataset.
		
		\subsection{Characteristics of the Dataset}
		
		In the past few years, many object detection datasets have been proposed in the field of remote sensing. However, most of them have some common disadvantages, for example, the number of instances and categories are relatively small, and the variations and diversity within the dataset are relatively limited, and the selection of categories does not satisfy the practical application. These shortcomings affect the development of object detection for remote sensing imagery to some extent.
		As far as we know, we provide the most diverse and challenging object dataset in comparison to other datasets in the field of remote sensing. There are 37 categories and more than a million of instances of objects in the FAIR1M dataset. We perform a comprehensive analysis of the FAIR1M dataset with other available object detection datasets, which is shown in Table \ref{table_dataset}. In addition to these analysis, the diversity and challenges of the FAIR1M dataset can be reflected in the following characteristics.

		\begin{enumerate}
			\itemsep=0pt
			\item{\textbf{\textit{Comprehensive fine-grained types.}}}
			To improve the development of fine-grained object detection and recognition, experts in remote sensing imagery interpretation are invited to design logical fine-grained types for the proposed dataset. The types we finally select are the most common categories in the practical applications. Taking the airplane as an example, researchers and practical applications mainly focus on the types of airplanes, whereas existing airplane datasets only consider coarse categories of airplane.

			\item {\textbf{\textit{Large range of sizes and orientations.}}} Due to the imaging principles and spatial resolutions of used sensors, it is common that multi-scale objects widely exist in remote sensing images. To extend the range of size variations of instances, we collect images with multiple spatial resolutions. In comparison with between-class size variation in other datasets, there are size variations and orientation variations not only between multiple classes but also within fine-grained types.
			
			\item{\textbf{\textit{High within-class variation and between-class similarity.}}} In addition to size variations and angle variations, high within-class variation and between-class similarity is one of the important characteristics of the FAIR1M dataset. For each group of objects, we select common fine-grained types to be different categories. However, the shapes and appearance of different fine-grained types are similar, which results in high between-class similarity. To obtain high within-class variation, we collect the same scenes with different seasons and weather conditions. Therefore, objects of the same category have different poses, background, colors and light. High within-class variation and between-class similarity make the FAIR1M dataset become more diverse and challenging.
			
			\item {\textbf{\textit{Complex scenes with densely distributed objects.}}} It is common that objects might be densely distributed in remote sensing images. To develop algorithms for densely distributed object detection in remote sensing images, we collect many images from complex scenes with densely distributed objects, such as large ports, parking lots, and crossroads, and we annotate the instances in the dense scenes one by one. Object detection in these scenes is a relatively enormous challenge for many of the current object detection algorithms.
			
			\item {\textbf{\textit{Rich geographic information.}}} The images in existing datasets in the field of remote sensing only contain the size information. Compared with these datasets, the FAIR1M dataset contains rich geographic information, such as spatial resolution, longitude, and latitude. Geographic information can provide researchers with more remote sensing information. In addition to the spatial information of the image, image scenes in the FAIR1M dataset are multi-temporal, which can be used to analyze the changes in the temporal dimension. 
			

		\end{enumerate}

		

		\begin{figure}
			\begin{minipage}{0.49\linewidth}
				\centerline{\includegraphics[width=1\linewidth]{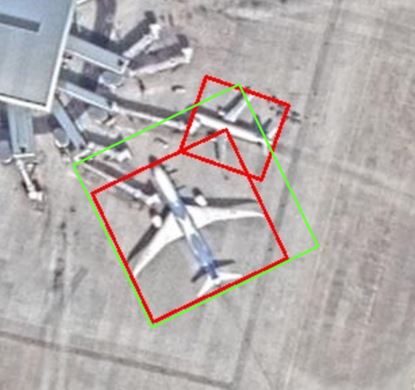}}
				\centerline{(a)}
			\end{minipage}
			\hfill
			\begin{minipage}{0.49\linewidth}
				\centerline{\includegraphics[width=1\linewidth]{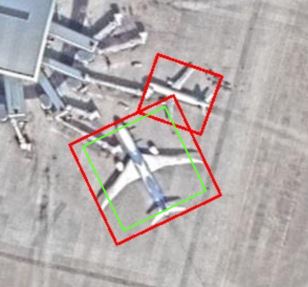}}
				\centerline{(b)}
			\end{minipage}
			\label{Figure:FAP1}
			\caption{{\color{black}{Two possible cases of prediction. (a) and (b) have the same IoU, but the predicted bounding box in (B) contains less confusing information. Green: prediction. Red: ground truth.}}}
		\end{figure}

		\begin{table*}[t]
			\renewcommand{\arraystretch}{1.2}
			\setlength{\tabcolsep}{0.8mm}
			\caption{The benchmark results of FAIR1M dataset on the task of oriented bounding boxes detection.}
			\label{table:result}

			\begin{tabular}{|c|c|c|c|c|c|c|c|c|c|c|c|}
				\hline
				\multicolumn{1}{|l|}{\multirow{2}{*}{Coarse Category}} & \multirow{2}{*}{Category} & \multicolumn{2}{c|}{\makecell[c]{RetinaNet\\\cite{lin2017focal}}} & \multicolumn{2}{c|}{\makecell[c]{Faster RCNN\\\cite{ren2015faster}}} & \multicolumn{2}{c|}{\makecell[c]{ROI Transformer\\\cite{ding2019learning}}} & \multicolumn{2}{c|}{\makecell[c]{Cascade RCNN\\\cite{cai2018cascade}}} & \multicolumn{2}{c|}{\makecell[c]{Gliding Vertex\\\cite{xu2020gliding}}} \\ \cline{3-12} 
				\multicolumn{1}{|l|}{}                                 &                           & $AP (\%)$             & $AP_{F} (\%)$           & $AP (\%)$             & $AP_{F} (\%)$            &$ AP (\%)$                & $AP_{F} (\%)$              & $AP (\%)$              & $AP_{F} (\%)$             & $AP (\%)$               & $AP_{F} (\%)$              \\ \hline
				\multirow{11}{*}{Airplane}                             & Boeing737                 & 38.46          & 13.75         & 36.43           & 10.10          & 39.58             & 17.60            & 40.42           & 12.53           & 35.43            & 11.54            \\ \cline{2-12} 
				& Boeing747                 & 55.36          & 20.37         & 50.68           & 11.81          & 73.56             & 35.53            & 52.86           & 23.37           & 47.88            & 13.42            \\ \cline{2-12} 
				& Boeing777                 & 24.75          & 5.62          & 22.50           & 3.20           & 18.32             & 7.51             & 29.07           & 11.06           & 15.67            & 3.32             \\ \cline{2-12} 
				& Boeing787                 & 51.81          & 13.57         & 51.86           & 13.42          & 56.43             & 31.22            & 52.47           & 21.77           & 48.32            & 11.23            \\ \cline{2-12} 
				& C919                      & 0.81           & 0.31          & 0.01            & 0.00           & 0.00              & 0.00             & 0.00            & 0.00            & 0.01             & 0.01             \\ \cline{2-12} 
				& A220                      & 40.50          & 14.33         & 47.81           & 21.88          & 47.67             & 30.21            & 44.37           & 20.69           & 40.11            & 12.33            \\ \cline{2-12} 
				& A321                      & 41.06          & 10.29         & 43.83           & 11.81          & 49.91             & 25.95            & 38.35           & 9.09            & 39.31            & 9.08             \\ \cline{2-12} 
				& A330                      & 18.02          & 4.73          & 17.66           & 2.25           & 27.64             & 13.98            & 26.55           & 10.89           & 16.54            & 2.21             \\ \cline{2-12} 
				& A350                      & 19.94          & 5.54          & 19.95           & 4.17           & 31.79             & 14.15            & 17.54           & 2.10            & 16.56            & 3.23             \\ \cline{2-12} 
				& ARJ21                     & 1.70           & 0.27          & 0.13            & 0.02           & 0.00              & 0.00             & 0.00            & 0.00            & 0.01             & 0.01             \\ \cline{2-12} 
				& other-airplane            & 62.75          & 28.11         & 66.15           & 19.29          & 68.28             & 35.51            & 65.64           & 30.83           & 61.04            & 20.11            \\ \hline
				\multirow{9}{*}{Ship}                                  & Passenger Ship            & 9.57           & 4.38          & 9.81            & 5.67           & 14.31             & 10.00            & 12.10           & 9.09            & 9.12             & 4.56             \\ \cline{2-12} 
				& Motorboat                 & 22.55          & 9.09          & 28.78           & 9.09           & 28.07             & 16.10            & 28.84           & 12.76           & 23.34            & 9.06             \\ \cline{2-12} 
				& Fishing Boat              & 1.33           & 0.39          & 1.77            & 1.01           & 1.03              & 1.00             & 0.71            & 0.36            & 1.23             & 0.34             \\ \cline{2-12} 
				& Tugboat                   & 16.37          & 9.09          & 17.65           & 9.09           & 14.32             & 9.09             & 15.35           & 9.09            & 15.67            & 9.09             \\ \cline{2-12} 
				& Engineering Ship          & 19.11          & 9.09          & 16.47           & 3.46           & 15.97             & 11.09            & 18.53           & 10.26           & 15.43            & 7.76             \\ \cline{2-12} 
				& Liquid Cargo Ship         & 14.26          & 9.09          & 16.19           & 9.09           & 18.04             & 10.99            & 14.63           & 9.09            & 15.32            & 8.98             \\ \cline{2-12} 
				& Dry Cargo Ship            & 24.70          & 9.70          & 27.06           & 9.09           & 26.02             & 13.84            & 25.15           & 13.99           & 25.43            & 9.34             \\ \cline{2-12} 
				& Warship                   & 15.37          & 9.09          & 13.16           & 6.67           & 12.97             & 7.77             & 14.53           & 9.09            & 13.56            & 8.56             \\ \cline{2-12} 
				& other-ship                & 2.63           & 0.77          & 3.04            & 2.02           & 2.25              & 1.66             & 1.54            & 0.55            & 2.45             & 1.23             \\ \hline
				\multirow{10}{*}{Vehicle}                              & Small Car                 & 65.20          & 15.85         & 68.42           & 21.60          & 68.80             & 38.95            & 68.19           & 37.54           & 66.23            & 16.67            \\ \cline{2-12} 
				& Bus                       & 22.42          & 9.09          & 28.37           & 9.09           & 37.41             & 14.29            & 28.25           & 8.66            & 23.43            & 8.45             \\ \cline{2-12} 
				& Cargo Truck               & 44.17          & 9.09          & 51.24           & 9.09           & 53.96             & 21.79            & 48.62           & 18.38           & 46.78            & 10.32            \\ \cline{2-12} 
				& Dump Truck                & 35.37          & 9.37          & 43.60           & 9.09           & 45.68             & 18.62            & 40.40           & 13.79           & 36.56            & 9.67             \\ \cline{2-12} 
				& Van                       & 52.44          & 13.19         & 57.51           & 13.82          & 58.39             & 26.55            & 58.00           & 20.96           & 53.78            & 13.45            \\ \cline{2-12} 
				& Trailer                   & 19.17          & 9.09          & 15.03           & 2.88           & 16.22             & 5.59             & 13.66           & 5.33            & 14.32            & 4.56             \\ \cline{2-12} 
				& Tractor                   & 1.28           & 0.85          & 3.04            & 2.34           & 5.13              & 6.90             & 0.91            & 0.79            & 16.39            & 1.45             \\ \cline{2-12} 
				& Excavator                 & 17.03          & 9.09          & 17.99           & 7.55           & 22.17             & 10.39            & 16.45           & 9.09            & 16.92            & 8.23             \\ \cline{2-12} 
				& Truck Tractor             & 28.98          & 9.09          & 29.36           & 9.09           & 46.71             & 23.62            & 30.27           & 9.09            & 28.91            & 10.21            \\ \cline{2-12} 
				& other-vehicle             & 8.91           & 4.22          & 5.23            & 1.45           & 11.62             & 9.09             & 11.65           & 9.09            & 8.98             & 4.58             \\ \hline
				\multirow{4}{*}{Court}                                 & Basketball Court          & 50.58          & 9.09          & 58.26           & 9.09           & 54.84             & 32.02            & 38.81           & 16.35           & 48.41            & 8.79             \\ \cline{2-12} 
				& Tennis Court              & 81.09          & 18.09         & 82.67           & 9.09           & 80.35             & 71.34            & 80.29           & 53.55           & 80.31            & 32.46            \\ \cline{2-12} 
				& Football Field            & 52.50          & 9.09          & 54.50           & 9.09           & 56.68             & 30.37            & 48.21           & 21.49           & 53.46            & 11.34            \\ \cline{2-12} 
				& Baseball Field            & 66.76          & 9.09          & 71.71           & 8.30           & 69.07             & 51.68            & 67.90           & 42.70           & 66.93            & 11.59            \\ \hline
				\multirow{3}{*}{Road}                                  & Intersection              & 60.13          & 9.09          & 59.86           & 9.09           & 58.44             & 31.18            & 55.67           & 20.87           & 59.41            & 8.41             \\ \cline{2-12} 
				& Roundabout                & 17.41          & 5.77          & 16.92           & 4.54           & 18.58             & 6.50             & 20.35           & 9.09            & 16.25            & 4.29             \\ \cline{2-12} 
				& Bridge                    & 12.58          & 6.06          & 11.87           & 4.77           & 31.81             & 15.42            & 12.62           & 7.88            & 10.39            & 5.15             \\ \hline
				\multicolumn{2}{|c|}{$mAP$/$mAP_{F}$ (\%)}                                                     & 30.19          & 8.99          & 31.53           & 7.92           & 34.65             & 19.12            & 30.78           & 14.09           & 29.46            & 8.51             \\ \hline
			\end{tabular}
			
		\end{table*}

		\subsection{Dataset Splits}
		The proposed dataset will be available in the ISPRS benchmark website. The proportions of the three subsets
		are 1/2, 1/6 and 1/3, respectively. We will provide all the images and ground truth for the training set and the validation set. For the testing set, we only provide images, but researchers can obtain detection results via the ISPRS benchmark evaluation server.

		\section{Algorithm Analysis}
		
		We consider two detection tasks and a recognition task on the FAIR1M dataset. The detection tasks are fine-grained oriented object detection and horizontal object detection in remote sensing images. In these object detection tasks, we need to detect objects belonging to the 37 categories in each image. Based on this task, we propose a cascaded hierarchical object detection method to improve the accuracy of fine-grained object detection. The other task is fine-grained visual categorization in remote sensing images. In this task, we set three fine-grained classification subtasks based on the FAIR1M dataset.
		
		\subsection{Detection}
		Object detection is a fundamental task in the field of remote sensing imagery. However, fine-grained categories, various sizes and orientations, multiple scenes severely impact the detection performance on the FAIR1M dataset. We have benchmarked several state-of-the-art object detection methods and our novel hierarchical object detection method.
		
		\begin{algorithm}[t]
			\KwData{a dataset of $C$ categories with ground truth boxes $G_{}^{c}$ and predicted bounding boxes $D_{}^{c}$}
			\KwResult{$mAP_{F}$}

			\For{pairs of ($G_{i}^{c}$, $D_{j}^{c}$)}{
				calculate $FIoU$ to determine $TP$, $FP$, $FN$\;
			}
			
			\For{C categories} {
				\For {scores} {
					$Precision_{F}$=$\sum_{i}^{Num_{G}}\frac{FIoU \cdot TP_{i} \cdot score_{TP}}{TP_{i} \cdot score_{TP}+FP_{i} \cdot score_{FP}}$ \\
					
					$Recall_{F}$=$Recall$=$\frac{FIoU \cdot TP}{TP+FN}$ \\
					
				}
			}
			calculate $mAP_{F}$ by obtained $Precision_{F}$ and $Recall_{F}$\;
			\caption{The calculation process of $mAP_{F}$}
		\end{algorithm}

		\begin{table}[]
			\renewcommand{\arraystretch}{1.3}
			\setlength{\tabcolsep}{1.5mm}
			\caption{The benchmark results of FAIR1M dataset on the task of horizontal bounding boxes detection.}
			\label{table:resulthbb}
			\begin{tabular}{|c|c|c|}
				\hline
				Method       & $mAP$   & $mAP_{F}$  \\ \hline
				RetinaNet \cite{lin2017focal}    & 30.68 & 8.45  \\ \hline
				Faster RCNN \cite{ren2015faster} & 32.32 & 16.13 \\ \hline
				Cascade RCNN \cite{cai2018cascade} & 34.12 & 16.57 \\ \hline
			\end{tabular}
		\end{table}

		\subsubsection{Evaluation Metrics}
		
		The evaluation will be based on the quantitative comparison between predictions and ground-truth for the dataset. In this paper, we design Fine-grained Intersection-Over-Union (FIoU) and Fine-grained mean Average Precision ($mAP_{F}$) for the FAIR1M dataset. These metrics are detailed in following sections.
		
		\textbf{Fine-grained Intersection-Over-Union (FIoU).}
		For the generic object detection task, each detected bounding box is labeled by the Intersection-Over-Union (IoU) metric with ground-truth for a specific category. The ratio of the area of the intersection to the union is the IoU metric of each pair of detected bounding box (D) and ground-truth (G). If the IoU is higher than the threshold of 0.5, the matched detection bounding box will be labeled as $TP$. Otherwise, it will be labeled as $FP$. If a ground-truth bounding box has no $TP$ detection to match with, it will be labeled as $FN$. The IoU metric can represent the overlap of two bounding boxes, but it ignores the impact of confusing information inside or outside the detection boxes. As shown in Figure 8, the ground truth box and predicted box have the same IoU in (a) and (b). However, (a) contains more information about another object and background, which is bad for fine-grained classification. Therefore, we design a novel FIoU for the task of fine-grained object detection to penalize the exceptional results.
		
		\begin{equation}
			\label{eq_precision}
			FIoU = \sqrt[3]{\frac{G \cap D}{G \cup D} \cdot \frac{G \cap D}{G} \cdot \frac{G \cap D}{D}}
		\end{equation}

		\begin{figure*}
			\centering
			\includegraphics[scale=0.48]{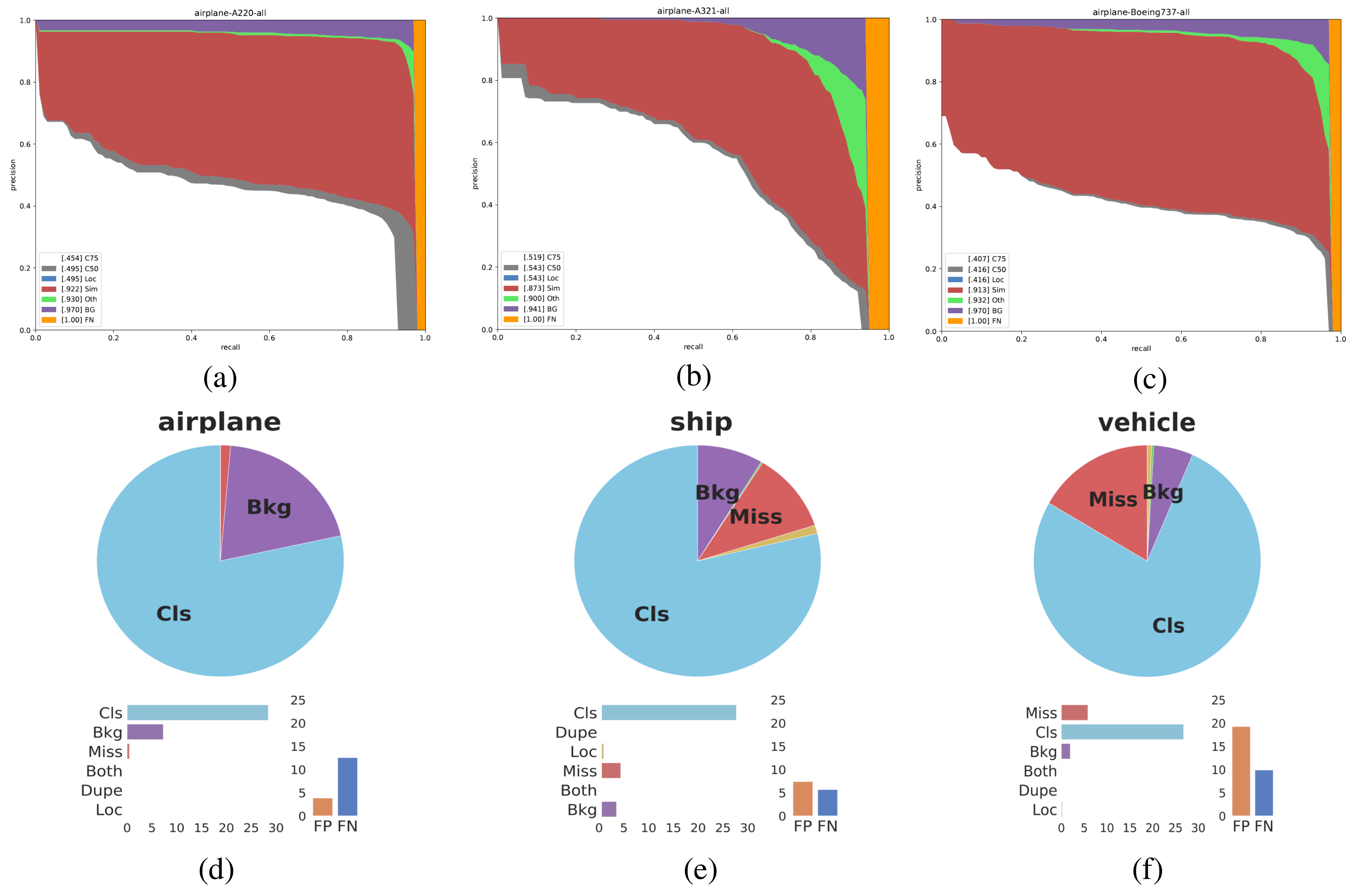}
			\caption{The false analysis of some categories in the FAIR dataset. (a), (b) and (c) show the false analyses between fine-grained airplanes using the COCO evalution tookit. (d), (e) and (f) show the false analyses of airplanes, ships and vehicles using TIDE toolbox.}
			\label{FIG:coco}
		\end{figure*}


		

		

		\begin{figure*}
			\begin{minipage}{0.49\linewidth}
				\centerline{\includegraphics[width=1\linewidth]{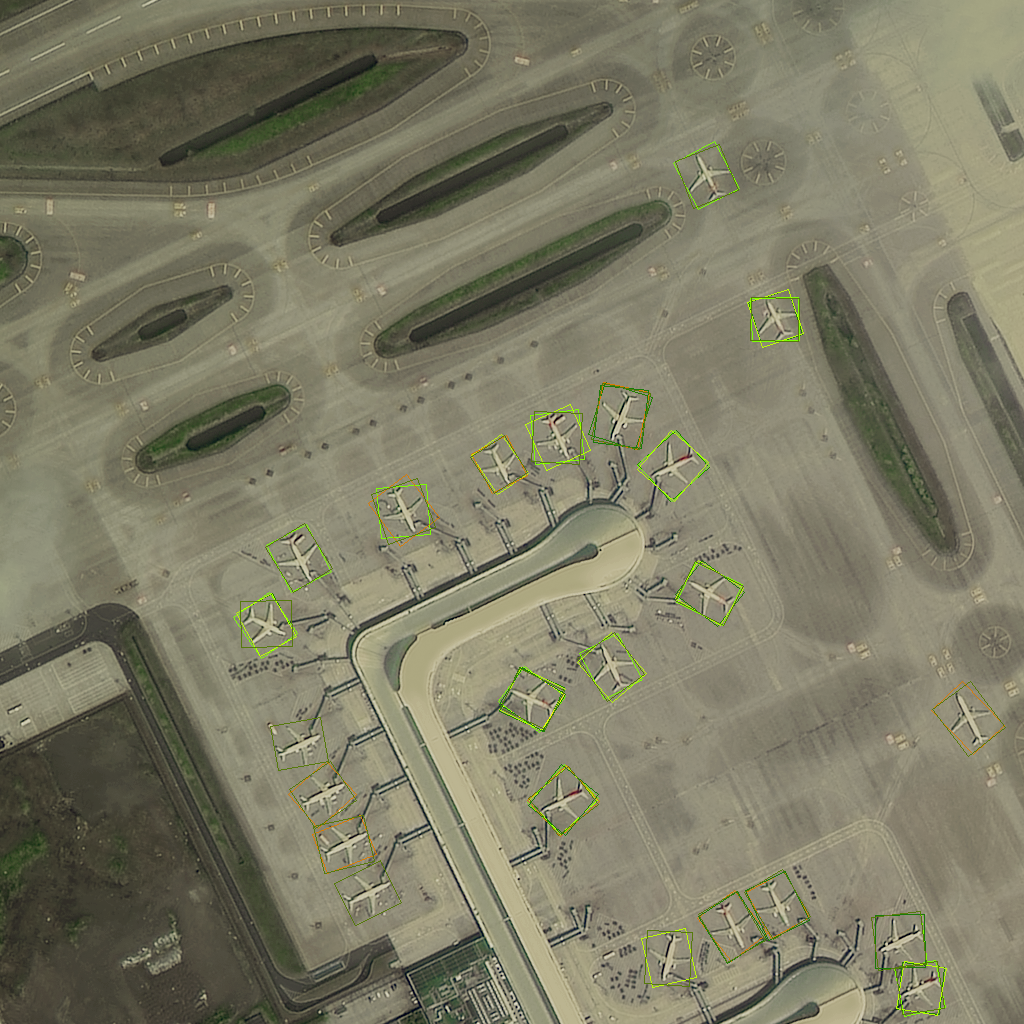}}
				\centerline{(a)}
			\end{minipage}
			\hfill
			\begin{minipage}{0.49\linewidth}
				\centerline{\includegraphics[width=1\linewidth]{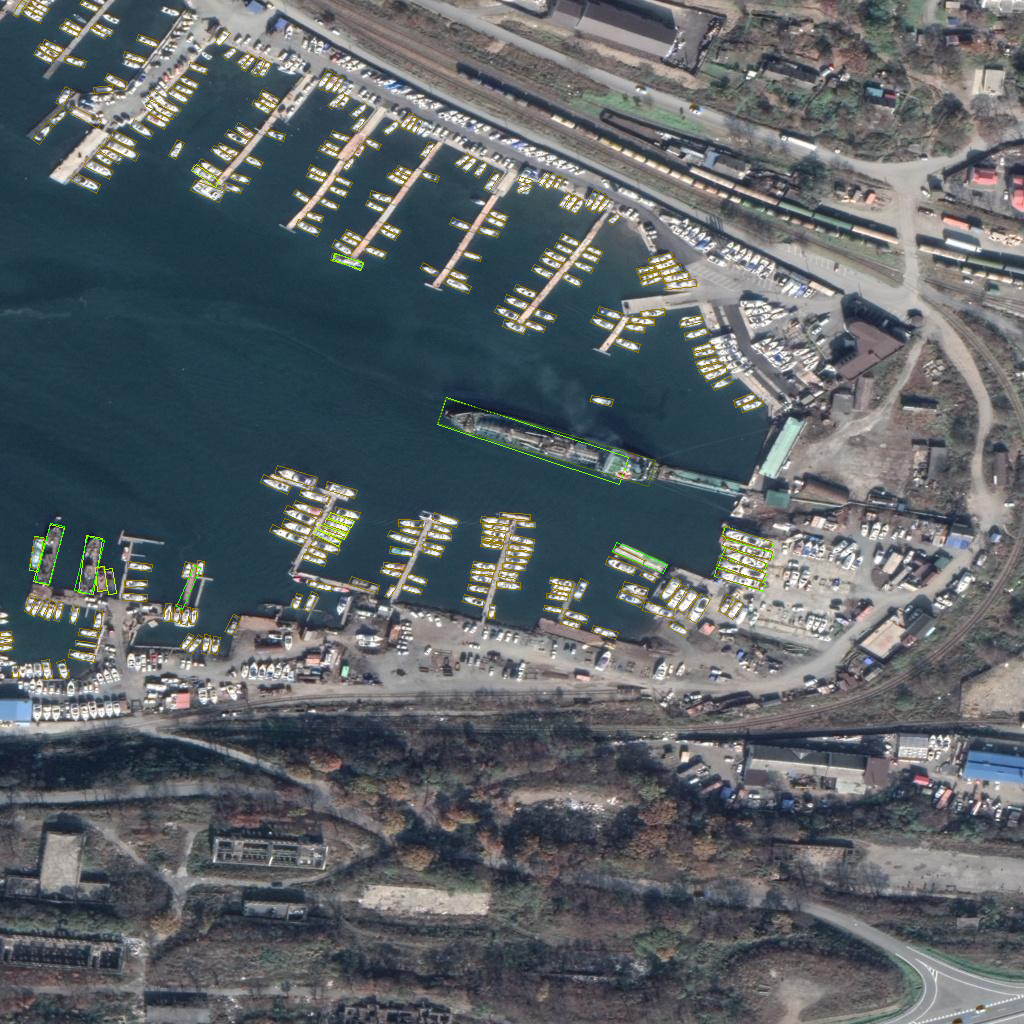}}
				\centerline{(b)}
			\end{minipage}
			\hfill
			\begin{minipage}{0.49\linewidth}
				\centerline{\includegraphics[width=1\linewidth]{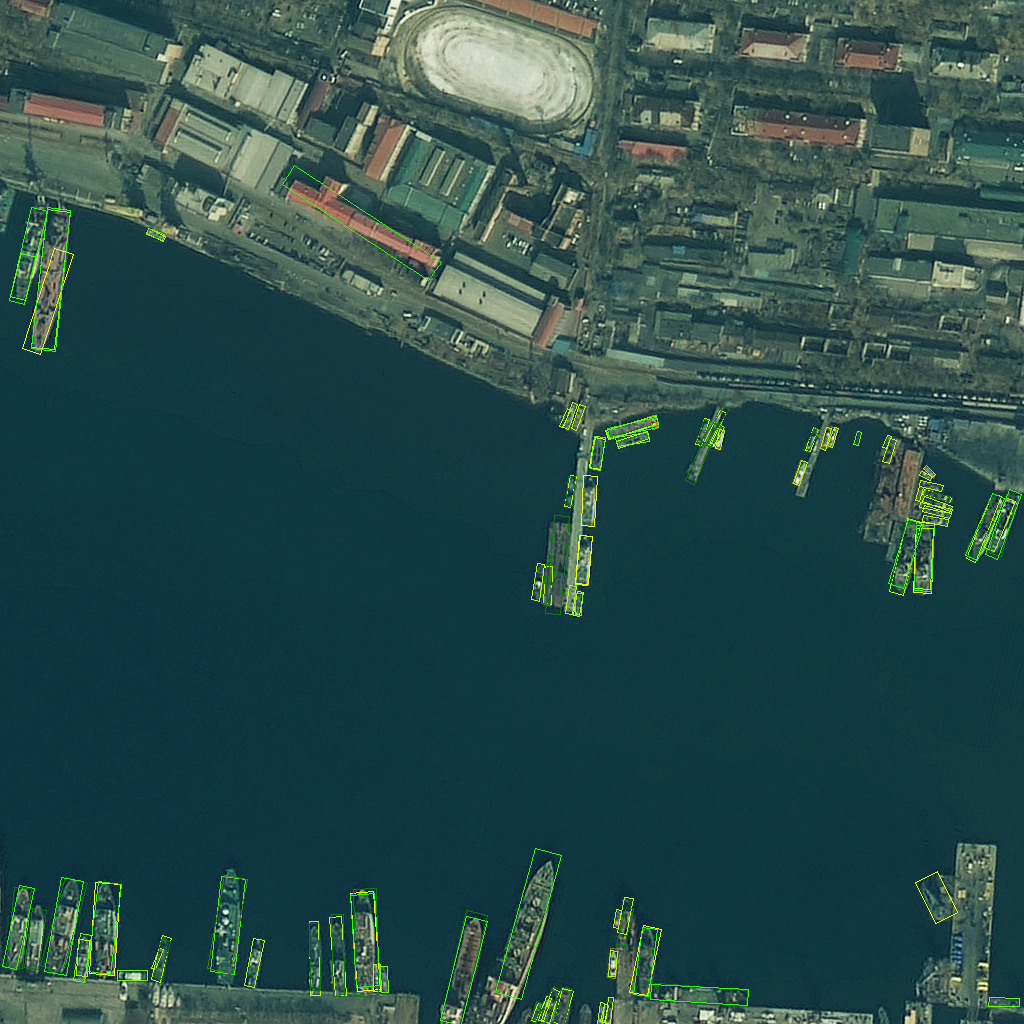}}
				\centerline{(c)}
			\end{minipage}
			\hfill
			\begin{minipage}{0.49\linewidth}
				\centerline{\includegraphics[width=1\linewidth]{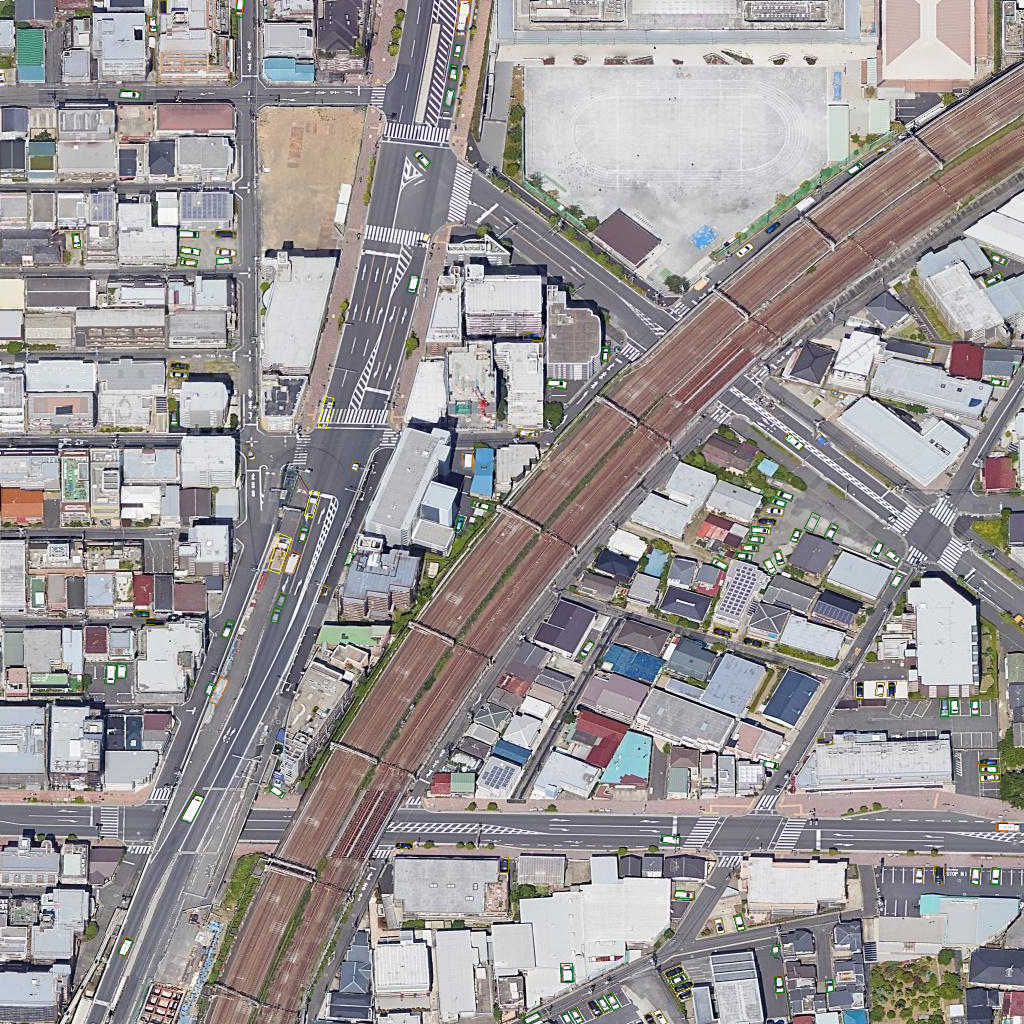}}
				\centerline{(d)}
			\end{minipage}
			\caption{{\color{black}{Visualization of detection results of testing on FAIR1M dataset using ROI Tranformer method.}}}
			\label{Figure: error}
		\end{figure*}

		\begin{figure*}[t]
			\centering
			\includegraphics[width=1.1\linewidth]{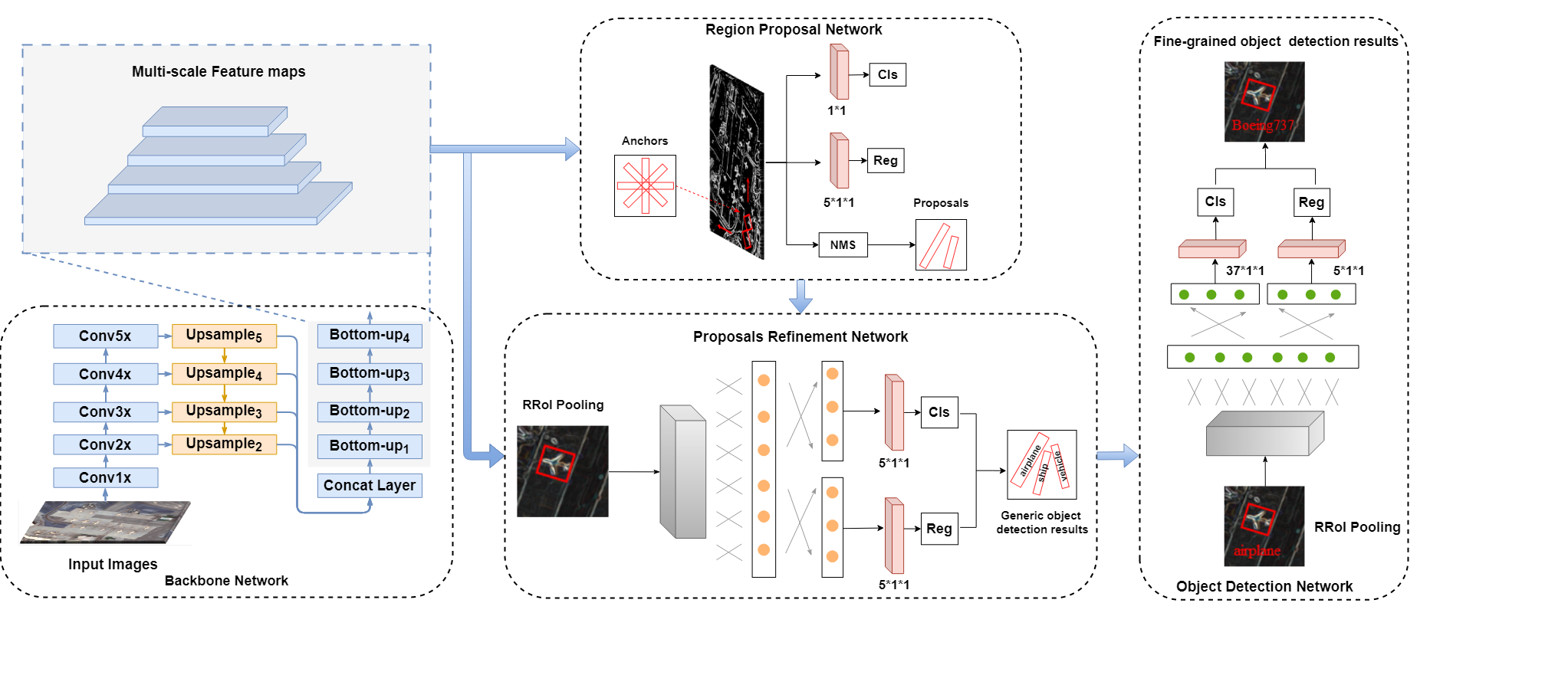}
			\caption{The structure of cascaded hierarchical object detection network.}
			\label{FIG:CHODN}
		\end{figure*}

		\textbf{Fine-grained mean Average Precision ($mAP_{F}$).}
		Different from generic object detection in remote sensing images, the task of fine-grained object detection pays more attention to type recognition. As a result, we use $FIoU$ to obtain $TP$ and $FP$. We define a detection box as  $TP$ if $FIoU$ is more than 0.5, otherwise it is $FP$. According to $TP$ and $FP$, we can calculate recall and precision. We add the classification score as a constraint to the original formula of the precision. High scores indicate that we obtain a better detector.
		\begin{equation}
			\label{eq_precision}
			Precision_{F}=\frac{FIoU \cdot TP \cdot score_{TP}}{TP \cdot score_{TP}+FP \cdot score_{FP}}
		\end{equation}
		
		\begin{equation}
			\label{eq_recall}
			Recall_{F}=Recall=\frac{FIoU \cdot TP}{TP+FN}
		\end{equation}
		where $score$ means the classification score of detected boxes. For each category, the precision is calculated with $TP$ and $FP$ as shown in Equation\ref{eq_precision}, under the $FIoU$ threshold of 0.5 and a series of score thresholds from the VOC2012 \citep{pascal-voc-2012}. Then the Average Precision ($AP_{F}$) is the mean of the collection of precisions. Finally, the $mAP_{F}$ is calculated with the mean of $AP_{F}$ over all categories. The $mAP_{F}$ is a float value between 0.0 and 1.0. Compared with $mAP$ in the task of generic object detection, $mAP_{F}$ is more sensitive to fine-grained classification scores.


		

		\begin{figure*}
			\begin{minipage}{0.49\linewidth}
				\centerline{\includegraphics[width=1\linewidth]{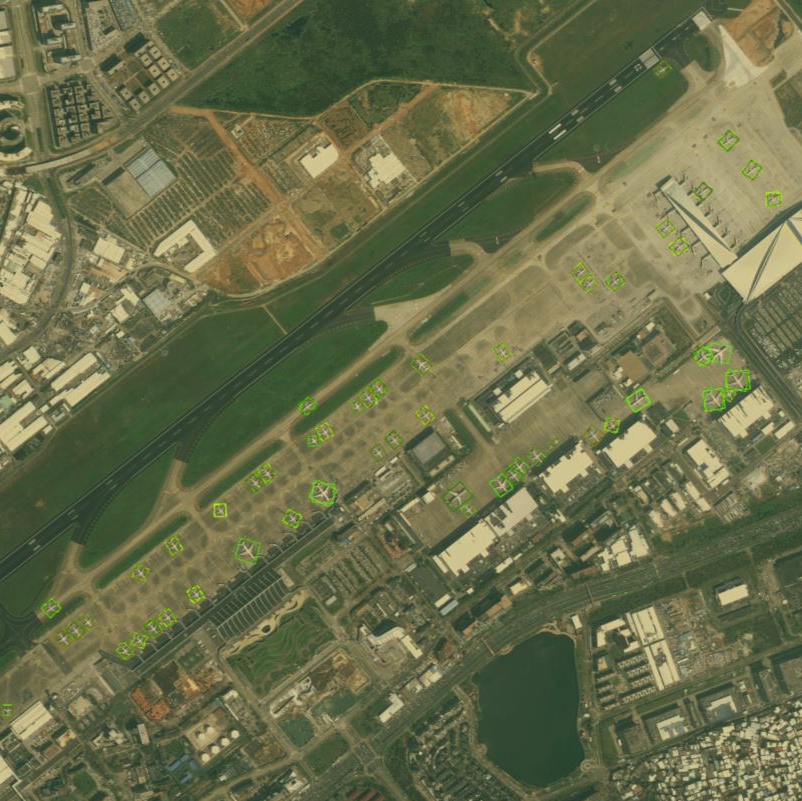}}
				\centerline{(a)}
			\end{minipage}
			\hfill
			\begin{minipage}{0.49\linewidth}
				\centerline{\includegraphics[width=1\linewidth]{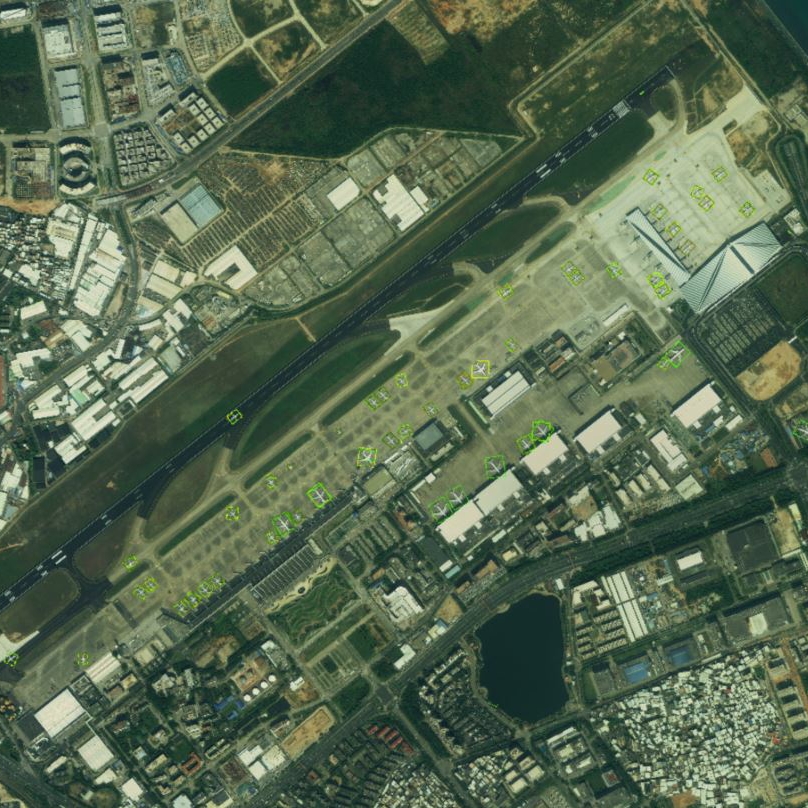}}
				\centerline{(b)}
			\end{minipage}
			\hfill
			\begin{minipage}{0.49\linewidth}
				\centerline{\includegraphics[width=1\linewidth]{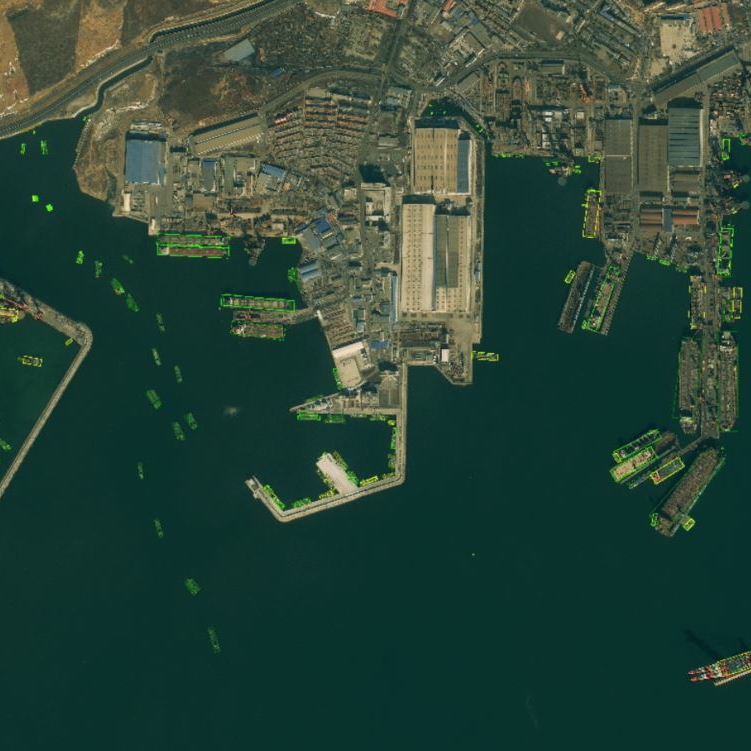}}
				\centerline{(c)}
			\end{minipage}
			\hfill
			\begin{minipage}{0.49\linewidth}
				\centerline{\includegraphics[width=1\linewidth]{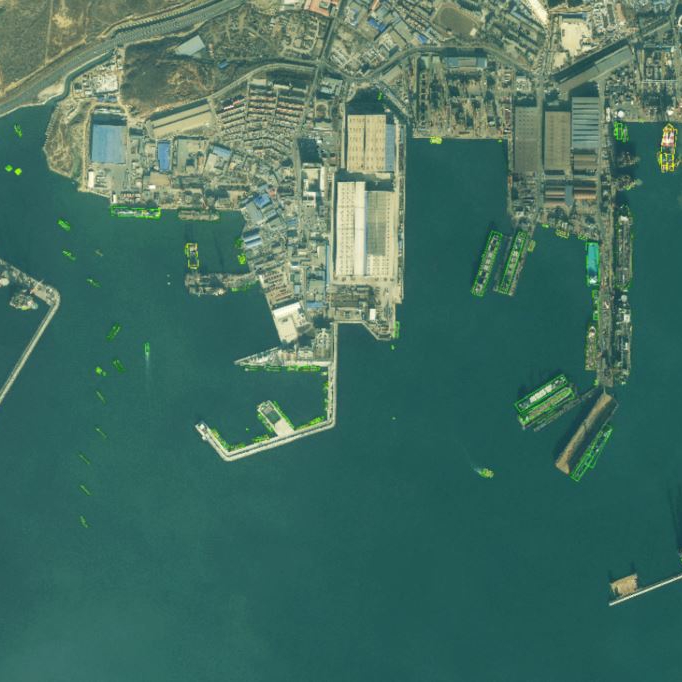}}
				\centerline{(d)}
			\end{minipage}
			\caption{{\color{black}{The detection results of CHODNet on multi-temporal images. (a), (b) Gaoqi airport. (c), (d) Dalian port.}}}
			\label{Figure: temporal}
		\end{figure*}

		\textbf{False analysis.}
		In addition to quantitative metrics, we conduct two detailed false analyses: COCO evaluation toolkit \cite{lin2014microsoft} and the TIDE toolbox \cite{bolya2020tide}. The COCO false analysis contains the curves of $C75$, $C50$, $Loc$, $Sim$, $Oth$, $BG$, and $FN$. We set the super-categories of all sub-categories to the corresponding generic categories. For example, the super-categories of Boeing737 and Boeing747 both are airplane. In this way, we can obtain the influence of similar objects and dissimilar objects. Compared with the COCO evaluation toolkit, the TIDE toolbox can calculate the contribution of each error type.
		

		
		
		
		\subsubsection{Benchmarks}
		
		\textbf{Dataset.} The sizes of images in the FAIR1M dataset are so large that images are not directly input to the object detection networks. In this paper, we crop images into 1024 $\times$ 1024 patches with a stride of 256.
		
		\textbf{Baseline models.}
		We have investigated the state-of-the-art oriented object detection algorithms in the field of object detection from remote sensing imagery. We select RetinaNet \cite{lin2017focal}, Faster R-CNN \cite{ren2015faster}, Cascade R-CNN \cite{cai2018cascade}, Gliding Vertex \cite{xu2020gliding}, and ROI Transformer \cite{ding2019learning} as our baseline models with a ResNet-101 backbone \cite{he2016deep}. The implement of these methods is built based on the code library \cite{chen2019mmdetection, ding2021object, xu2020gliding}. To be specific, RetinaNet, Faster R-CNN and Cascade R-CNN represent region-based methods and proposal-based methods respectively, which are transferred from horizontal object detection in natural scenes. ROI Transformer and Gliding Vertex represent different oriented object detection methods in the field of remote sensing.

		
		\textbf{Baseline results.}
		We train and evaluate 5 detectors on the task of oriented bounding boxes detection and 3 detectors on the task of horizontal bounding boxes detection. The results in Table \ref{table:result} and Table \ref{table:resulthbb} show the difficulty of generic object detection methods in detecting fine-grained objects. We calculate $AP$ and $AP_{F}$ for each category. For the categories with obvious features, such as Boeing747, detectors obtain better results. The performance of detectors on most categories verifies the difficulty of our dataset. The number, distribution, and characteristics of different objects have caused the imbalance of the detection accuracy. Generally, objects with a large number of instances and obvious features are easier to obtain higher detection accuracy. While, some categories of object detection accuracy is quite low, for example, C919 and ARJ21. The main reason for this result is that these two types of airplanes are very rare, so it is necessary to study few-shot learning methods to improve the detection accuracy of these objects. Although the $mAP$ values of the five algorithms are close, there are great differences in the $mAP_{F}$ value. The combination of AP and $AP_{F}$ is helpful for analyzing the performance of the detectors. 
		
		We conduct false analysis on each category to further verify the challenge of our dataset. Figure \ref{FIG:coco} shows two kinds of false analyzes of several categories on the result of the ROI Transformer method. We plot $C75$, $C50$, $Loc$, $Sim$, $Oth$, $BG$, and $FN$ to analyze the within-class performance. Figure \ref{FIG:coco} (a), (b) and (c) show that:
		\begin{enumerate}
			\item Removing localization error ($Loc$) brings some improvement on $AP$, which shows the influence of the large range of sizes and orientations in our dataset.
			
			\item Removing the $Sim$ and $FN$ results in more improvements than removing the $Oth$ for each category, which means the influence of similar objects is larger than the influence of dissimilar objects for most categories. It is challenging that researchers need to design more suitable algorithms to detect fine-grained objects accurately.
			
			\item In addition to $Sim$ and $FN$, complex background also has a large impact on detection results. The complex background in our dataset is also challenging.
			
		\end{enumerate}
		
		Figure \ref{FIG:coco} also shows the contribution of each error type for each category. It is obvious that the classification error accounts for most of the error. Moreover, missdetection has a great impact on ships and vehicles due to their small sizes. The results show more difficulty in detecting objects in the FAIR1M dataset.
		
		\textbf{Failure cases. }Figure \ref{Figure: error} shows the visualization of detection results using the ROI Transformer model. In addition to classification, there are undetected objects in the results. The complex background and small size of objects result in the misdetection of small objects.

		\subsubsection{Hierarchical Object Detection Method}
		
		According to the above experiments, although fine-grained detection is a detection task, the challenge lies in how to classify the objects. It is relatively difficult for detectors to learn the feature information of the 37 fine-grained categories. The categories of objects in the FAIR1M dataset are organized in a hierarchy, which is designed coarse to fine. Considering this characteristic of the FAIR1M dataset, we propose a cascaded hierarchical object detection network (CHODNet). It can learn external and internal representations independently from the dataset using a cascaded hierarchical structure. Figure \ref{FIG:CHODN} shows the structure of the  CHODNet.

		\textbf{Framework. }We build the CHODNet based on oriented Faster RCNN. Compared with Faster RCNN, CHODNet adds a training stage to learn the information of coarse categories. As shown in Figure \ref{FIG:CHODN}, CHODNet consists of four stages: feature refinement network, region proposal network, proposals refinement network, and fine-grained detection network.

		\begin{itemize}
			\item \textbf{Feature Refinement Network. }We use the ResNet-101 as our backbone.  As we know, the high-level layers and low-level layers in a deep learning network contain semantic information and localization information \cite{fu2020rotation}, respectively. To generate more discriminative feature maps, we upsample multi-scale feature maps to the same scale, and build a bottom-up feature hierarchy. 
			\item \textbf{Region Proposal Network.} Due to the oriented annotations in the FAIR1M dataset, we generate oriented anchors in the first stage. Each anchor can be represented as a five-tuple ($x$, $y$, $w$, $h$, $\theta$), where ($x$, $y$) denotes the center coordinates of the anchor, ($w$, $h$) specifies the width and height of it, respectively, and $\theta$ means the angle of it. We generate $K$ anchors with $K_{r}$ aspect ratios, $K_{s}$ scales, and $K_{a}$ angles, where $K=K_{r}$ $\times$ $K_{s}$ $\times$ $K_{a}$. At each position of the feature map, a regression output layer generates ($K$ $\times$ $5$) vectors to encode the offset of anchors, and a classification output layer generates ($K$ $\times$ $2$) scores to predict whether the anchor is positive. The loss function of this stage is a multi-task loss similar to the one of Faster RCNN, which is defined as:
			
			\begin{equation}
				L_{orpn} = L_{cls}(p,u) + L_{loc}(t,t^{*}).
			\end{equation}
			where $L_{cls}$ and $L_{loc}$ represent the classification loss and localization loss in the oriented region proposal network, respectively. For $L_{cls}$, the parameter $u$ is 1 if the anchor is positive, otherwise it is 0. The parameter $p$ denotes the predicted score of anchors on the background and foreground. $t$ and $t^{*}$ denote the predicted regression offset and ground-truth box, respectively.
			
			\item \textbf{Proposals Refinement Network. }Compared with the R-CNN subnetwork in the Faster RCNN, the proposal refinement network is mainly used to output the coarse category classification result and localization offset for each proposal. There are 5 coarse categories in the FAIR1M dataset. Therefore, the output layers of classification and regression both are of size $K$ $\times$ $5$. The loss function is still a multi-task loss about the classification and localization:
			
			\begin{equation}
				L_{prn} = L_{cls}(p,v) + L_{loc}(t, t^{*})
			\end{equation}
			where $v$ denotes the score of coarse categories.
			
			\item \textbf{Fine-grained Detection Network. }Different from the previous stage, the fine-grained detection network will output scores of all fine-grained categories. The parameters in fine-grained detection network are learned specifically for fine categories. As shown in Figure \ref{FIG:CHODN}, the output layers of classification and regression are $K$ $\times$ $37$ and $K$ $\times$ $5$, respectively. The loss of this stage is defined as:
			
			\begin{equation}
				L_{fdn} = L_{cls}(p,w) + L_{loc}(t,t^{*})
			\end{equation}
			Where, $w$ denotes the score of fine categories.
			
		\end{itemize}
		
		\textbf{Loss Function. }There are 3 stages in our CHODNet. The loss function of CHODNet is a weighted summation of oriented region proposal network, proposals refinement network, and fine-grained object detection network. In other words, the total loss consists of foreground/background loss, coarse loss, and fine-grained loss.
		
		\begin{equation}
			L_{CHODNet} = L_{orpn} + \lambda_{1}L_{prn} + \lambda_{2}L_{fdn}
		\end{equation}
		Where $\lambda_{1}$ and $\lambda_{2}$ denote the weights of different losses.
		
		\textbf{Staged Training Strategy. }We have defined a weighted summation of different losses. When training the detector, we hope it can give priority to learning the features of coarse categories. After the parameters of the current two stages are trained, we focus on the learning of the third stage. Therefore, we develop a stage training strategy to focus on different losses in an end-to-end manner. The staged training strategy can modify the value of the weights $\lambda_{1}$ and $\lambda_{2}$ while training the detector, which denote the contributions of different losses. For the three-stage detecting, the initial weights of the loss are [0.7, 0.3], then they are adjusted to [0.3, 0.7] after two hundred thousand steps. The loss function with the highest weight can be regarded as the focus of the current step.

		\textbf{Results and Temporal analysis. } After adopting the staged training strategy, CHODNet obtains 32.46\% mAP. Adding an extra branch of learning coarse categories can improve the accuracy of fine-grained object detection, which validates the effectiveness of hierarchical object detection framework. As shown in Figure \ref{Figure: temporal}, we test the CHODNet on multi-temporal images. The two rows of Figure \ref{Figure: temporal} show the Gaoqi airport and Dalian port at different times. Figure \ref{Figure: temporal} (a) and (b) are the Gaoqi Airport in 2015 and 2016, respectively. Figure \ref{Figure: temporal} (c) and Figure \ref{Figure: temporal} (d) are the Dalian port in 2016 and 2017, respectively. Although the lighting and timing of the images are different, the detectors can still detect the objects.


		
		\begin{figure*}
			\begin{minipage}{0.3\linewidth}
				\centerline{\includegraphics[width=1\linewidth]{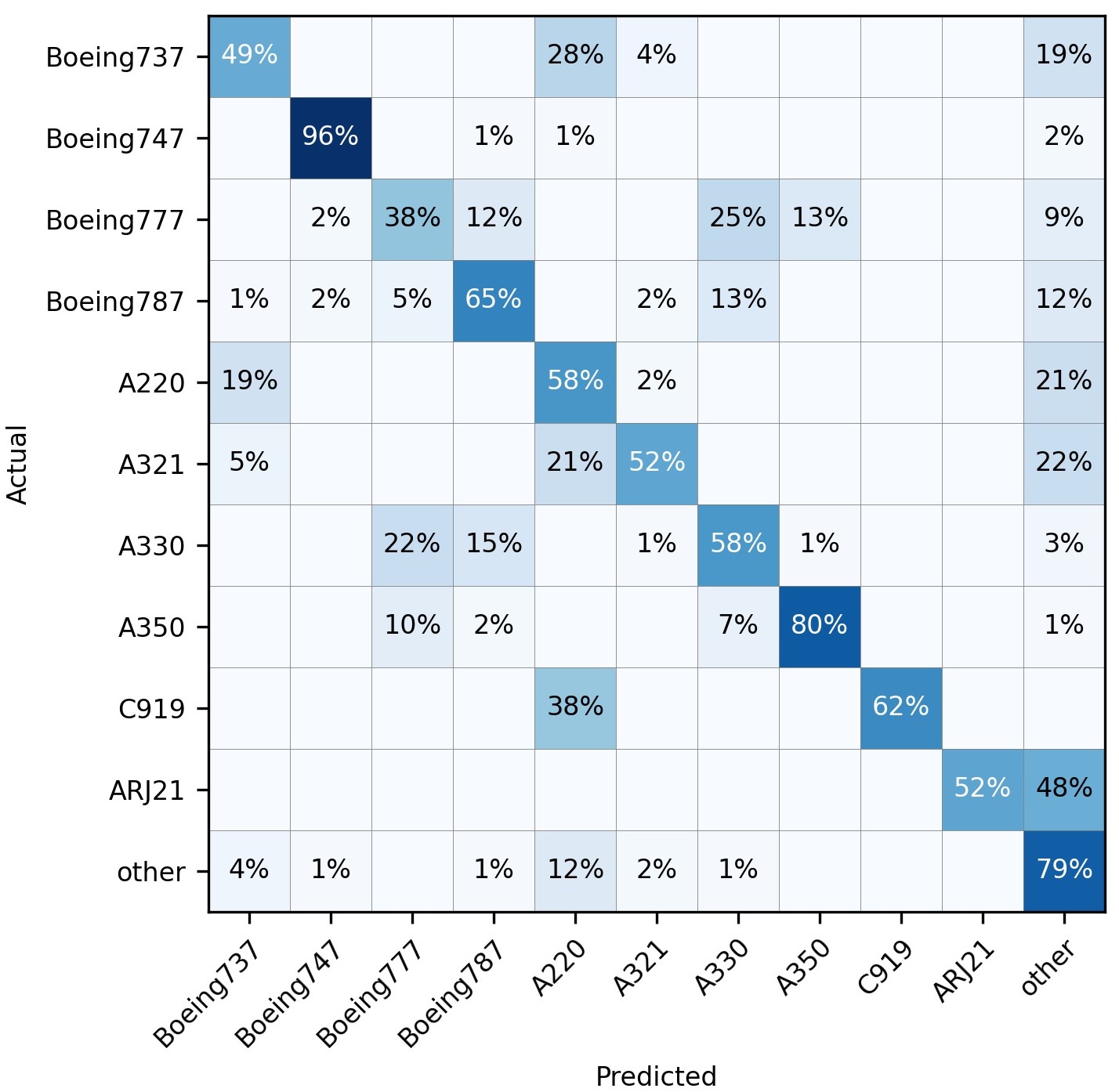}}
				\centerline{(a)}
			\end{minipage}
			\hfill
			\begin{minipage}{0.3\linewidth}
				\centerline{\includegraphics[width=1\linewidth]{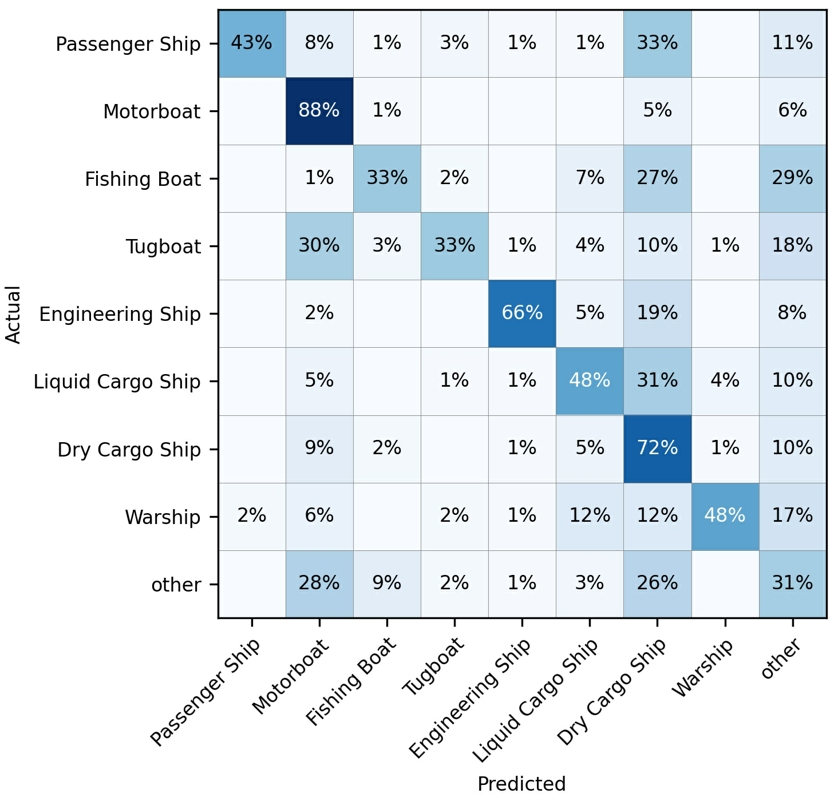}}
				\centerline{(b)}
			\end{minipage}
			\hfill
			\begin{minipage}{0.3\linewidth}
				\centerline{\includegraphics[width=1\linewidth]{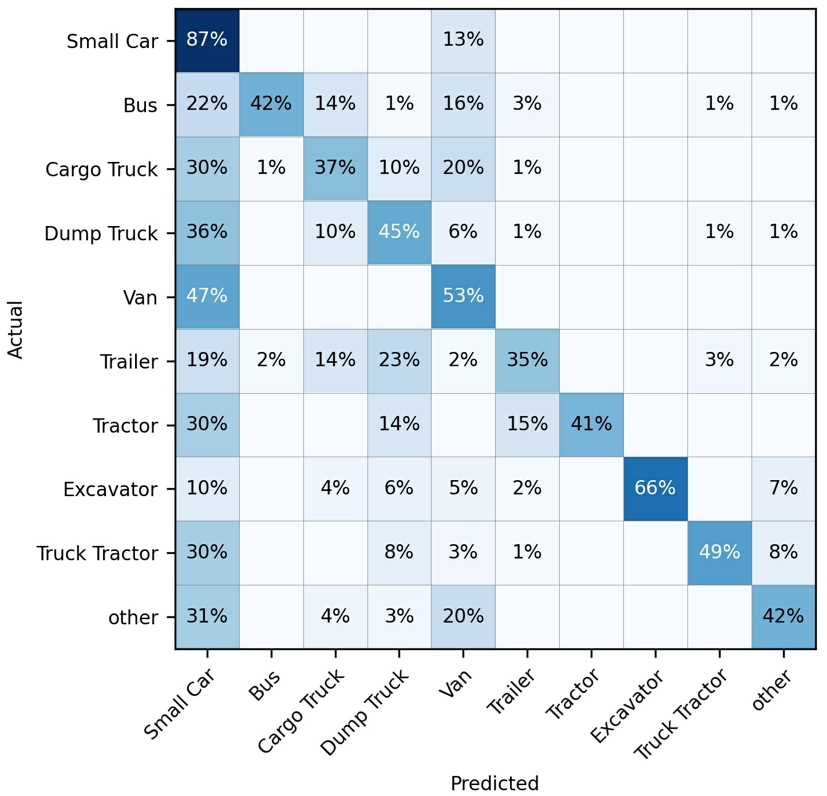}}
				\centerline{(b)}
			\end{minipage}
			\caption{{\color{black}{The confusion matrices of MMAL-Net on three fine-grained datasets.}}}
			\label{Figure: confusion}
		\end{figure*}

		\subsection{Fine-grained Image Classification}
		Fine-grained visual categorization (FGVC) aims to obtain the category of an input image. Many FGVC datasets have been proposed in the field of natural images, such as the CUB-200-2011 dataset \cite{WahCUB_200_2011}, the Stanford Cars dataset \cite{KrauseStarkDengFei-Fei_3DRR2013}, the FGVC-Aircraft dataset \cite{wu2014aircraft}, and Stanford Dogs dataset \cite{KhoslaYaoJayadevaprakashFeiFei_FGVC2011}. There are relatively few datasets about FGVC in the field of remote sensing imagery. HR-SAR and MR-SAR \cite{xu2020distribution} are two SAR ship datasets. Sumbul et al. \cite{sumbul2017fine} propose a street tree dataset in aerial images. It is necessary to build FGVC datasets and develop the FGVC task for movable objects in remote sensing images. Based on the FAIR1M dataset, we generate three FGVC datasets: FAIR-Airplane dataset, FAIR-Ship dataset, and FAIR-Vehicle dataset.
		
		\textbf{Dataset Information. } According to the bounding box annotations of the FAIR1M dataset, we select relatively larger airplanes, ships, and vehicles and crop them from the original images. The information about the number of instances of the three datasets can be seen in Table \ref{table:FGVC}.
		
		\begin{table}[]
			\renewcommand{\arraystretch}{1.5}
			\setlength{\tabcolsep}{3.5mm}
			\caption{The information of three FGVC datasets.}
			\label{table:FGVC}
			\begin{tabular}{|c|c|}
				\hline
				Category        & Instance \\ \hline
				Airplane\_train & 8000    \\ \hline
				Airplane\_test  & 4000     \\ \hline
				Ship\_train     & 8000    \\ \hline
				Ship\_test      & 4000    \\ \hline
				Vehicle\_train  & 10000    \\ \hline
				Vehicle\_test   & 6000    \\ \hline
			\end{tabular}
		\end{table}
		
		\begin{table}[]
			\renewcommand{\arraystretch}{1.5}
			\setlength{\tabcolsep}{3.5mm}
			\caption{The results of three FGVC methods on our datasets. The evaluation metric is $mAP$ and its provided values refer to \%.}
			\label{table:FGVC-result}
			\begin{tabular}{|c|c|c|c|}
				\hline
				Method     & Airplane & Ship  & Vehicle \\ \hline
				ResNet-50  & 41.23    & 37.86 & 38.93   \\ \hline
				ResNet-101 & 43.46    & 38.64 & 39.76   \\ \hline
				MMAL-Net   & 45.18    & 40.27 & 41.43   \\ \hline
			\end{tabular}
		\end{table}
		
		\textbf{Evaluation Metrics. }To evaluate the performance of FGVC algorithms, we use the classification accuracy as the metric for this task. $N_{correct}$ and $N_{all}$ denote the number of correctly predicted instances and total instances, respectively. The definition of the classification accuracy is as follows:
		
		\begin{center}
			\begin{equation}
				Acc = \frac{N_{correct}}{N_{all}}
			\end{equation}
		\end{center}
		
		\textbf{Baseline Classifiers and Results. }We choose two fundamental image classifiers (\textit{i.e. }ResNets) and a state-of-the-art fine-grained image classifier (\textit{i.e. }MMAL-Net \cite{zhang2020threebranch}) as our baseline classifiers. ResNet-50 and ResNet-101 are usually used for generic image classification. MMAL-Net has achieved state-of-the-art results on the CUB200-2011, FGVC-Aircraft and Stanford Cars datasets. 
		
		We train the three algorithms on three fine-grained datasets. As shown in Table \ref{table:FGVC-result}, ResNet-50 and ResNet-101 show more difficulty in fine-grained classification in remote sensing images. To show the results of various categories more clearly, we provide the confusion matrices on the three fine-grained datasets using the MMAL-Net method. Except for the small number of objects (\textit{i.e. } regarding the sub-categories \textit{C919} and \textit{ARJ21}), the MMAL-Net performs relatively well on the FAIR-Airplane dataset. However, the MMAL-Net shows more difficulty in classifying fine-grained ships and vehicles. Only \textit{Motorboat} and \textit{Small Car} have relatively high confidence. The remaining categories are confused with other categories to some extent. Therefore, it is challenging to design fine-grained classifiers on three datasets for researchers.

		\begin{table*}[]
			\renewcommand{\arraystretch}{1.5}
			\setlength{\tabcolsep}{3.65mm}
			\caption{Results of cross-dataset generalization. ROI Transformer-D and ROI Transformer-F are trained based on the DOTA dataset and FAIR1M dataset, respectively. The evaluation metric is $mAP$ and its provided values refer to \%.}
			\label{table:cross-result}
			
			\begin{tabular}{|c|c|c|c|c|c|}
				\hline
				Testing set             & Detector          & Airplane & Ship  & Vehicle & Aug.  \\ \hline
				\multirow{2}{*}{DOTA}   & ROI Transformer-D & 90.28    & 81.31 & 71.85   & 81.15 \\ \cline{2-6} 
				& ROI Transformer-F & 81.54    & 61.45 & 59.10   & 67.36 \\ \hline
				\multirow{2}{*}{FAIR1M} & ROI Transformer-D & 80.95    & 33.66 & 45.09   & 53.23 \\ \cline{2-6} 
				& ROI Transformer-F & 90.74    & 39.06 & 51.98   & 60.59 \\ \hline
			\end{tabular}
		\end{table*}
		
		\subsection{Cross-dataset Validations}
		In order to verify the generalization performance of our dataset, we use the DOTA dataset to do cross-dataset validation experiments, which is one of the largest and OBB-style object detection datasets in the field of remote sensing. For there are large-scale images in the DOTA dataset, we crop images into patches with 1024 $\times$ 1024 pixels. We only choose generic categories in the FAIR1M dataset to do cross-dataset validations because the DOTA dataset does not have fine-grained categories. ROI Transformer is used to be the testing detector for the experimental results presented in Table \ref{table:cross-result}. 
		
		The difference between the two datasets is 27.92\% mAP and 6.77\% mAP, respectively. The results in Table \ref{table:cross-result} show that the FAIR1M dataset covers the characteristics of the DOTA dataset and has more types and patterns not contained in the DOTA dataset. ROI Transformer-D and ROI Transformer-F get lower results on FAIR1M, which verifies the challenge of the FAIR1M dataset.
		

		\section{Conclusion}
		In this paper, we propose a more challenging dataset for fine-grained object detection and recognition in high-resolution remote sensing imagery. We believe the diversity and challenge of the FAIR1M dataset will benefit from fine-grained types, large range of sizes and orientations, high within-class variation and between-class similarity, complex scenes, and geographic information. We introduce the collection, category, annotation and characteristics of our dataset. Finally, we implement a series of state-of-the-art algorithms to build an object detection benchmark to foster future research. With the development of remote sensing image interpretation technology, coarse object recognition cannot meet the requirements well. We hope that this dataset can enhance the development of fine-grained object recognition in the field of remote sensing images.

		\section*{Acknowledgment}
		\label{acknowledgements}
		We are very grateful for the support of the ISPRS Scientific Initiatives 2021. To promote the academic research on object detection and recognition in high-resolution satellite images, the FAIR1M dataset will be a standard and large-scale dataset foundation for the ISPRS benchmark. It is also supported by the National Natural Science Foundation of China 61725105 and the Major project of China High-resolution Earth Observation System GFZX0404120201/GFZX0404120205.

		\printcredits
		
		\bibliographystyle{cas-model2-names}
		
		\bibliography{cas-dc-template}

		
	\end{sloppypar}
\end{document}